\documentclass[conference]{IEEEtran}
\usepackage{times}
\usepackage[numbers]{natbib}
\usepackage{multicol}
\usepackage[bookmarks=true]{hyperref}
\usepackage{graphics} 
\usepackage{epsfig} 
\usepackage{times} 
\usepackage{amsmath} 
\usepackage{amssymb}  
\usepackage{amsthm}
\usepackage{hyperref}
\usepackage{multirow}
\usepackage{caption,setspace}
\usepackage{booktabs}
\usepackage{arydshln}
\usepackage{tablefootnote}
\usepackage{csquotes}
\usepackage{footnote}
\makesavenoteenv{tabular}
\makesavenoteenv{table}
\usepackage[]{threeparttable}
\usepackage{breqn}
\usepackage{subcaption}
\usepackage{algorithm, algpseudocode}
\usepackage{float}
\usepackage[dvipsnames]{xcolor}
\let\labelindent\relax
\usepackage{enumitem}
\usepackage{tikz}

\usepackage{graphicx}  
\usepackage{subcaption}  

\DeclareMathOperator*{\argmin}{arg\,min}
\newtheorem{definition}{Definition}

\newtheorem{lemma}{Lemma}

\begin{document}

\title{Flow Matching Ergodic Coverage}


\author{%
  \IEEEauthorblockN{%
    Max Muchen Sun,\ 
    Allison Pinosky,\ 
    Todd Murphey%
  }%
  \IEEEauthorblockA{%
    Center for Robotics and Biosystems, Northwestern University, Evanston, IL 60208\\
    Email: msun@u.northwestern.edu\\
    Project website: \textcolor{blue}{\url{https://murpheylab.github.io/lqr-flow-matching/}}
  }%
}

\maketitle
\allowdisplaybreaks

\begin{abstract}
    Ergodic coverage effectively generates exploratory behaviors for embodied agents by aligning the spatial distribution of the agent's trajectory with a target distribution, where the difference between these two distributions is measured by the ergodic metric. However, existing ergodic coverage methods are constrained by the limited set of ergodic metrics available for control synthesis, fundamentally limiting their performance. In this work, we propose an alternative approach to ergodic coverage based on flow matching, a technique widely used in generative inference for efficient and scalable sampling. We formally derive the flow matching problem for ergodic coverage and show that it is equivalent to a linear quadratic regulator problem with a closed-form solution. Our formulation enables alternative ergodic metrics from generative inference that overcome the limitations of existing ones. These metrics were previously infeasible for control synthesis but can now be supported with no computational overhead. Specifically, flow matching with the Stein variational gradient flow enables control synthesis directly over the score function of the target distribution, improving robustness to the unnormalized distributions; on the other hand, flow matching with the Sinkhorn divergence flow enables an optimal transport-based ergodic metric, improving coverage performance on non-smooth distributions with irregular supports. We validate the improved performance and competitive computational efficiency of our method through comprehensive numerical benchmarks and across different nonlinear dynamics. We further demonstrate the practicality of our method through a series of drawing and erasing tasks on a Franka robot.
\end{abstract}

\IEEEpeerreviewmaketitle

\begin{figure*}
    \centering
    \includegraphics[width=0.99\textwidth]{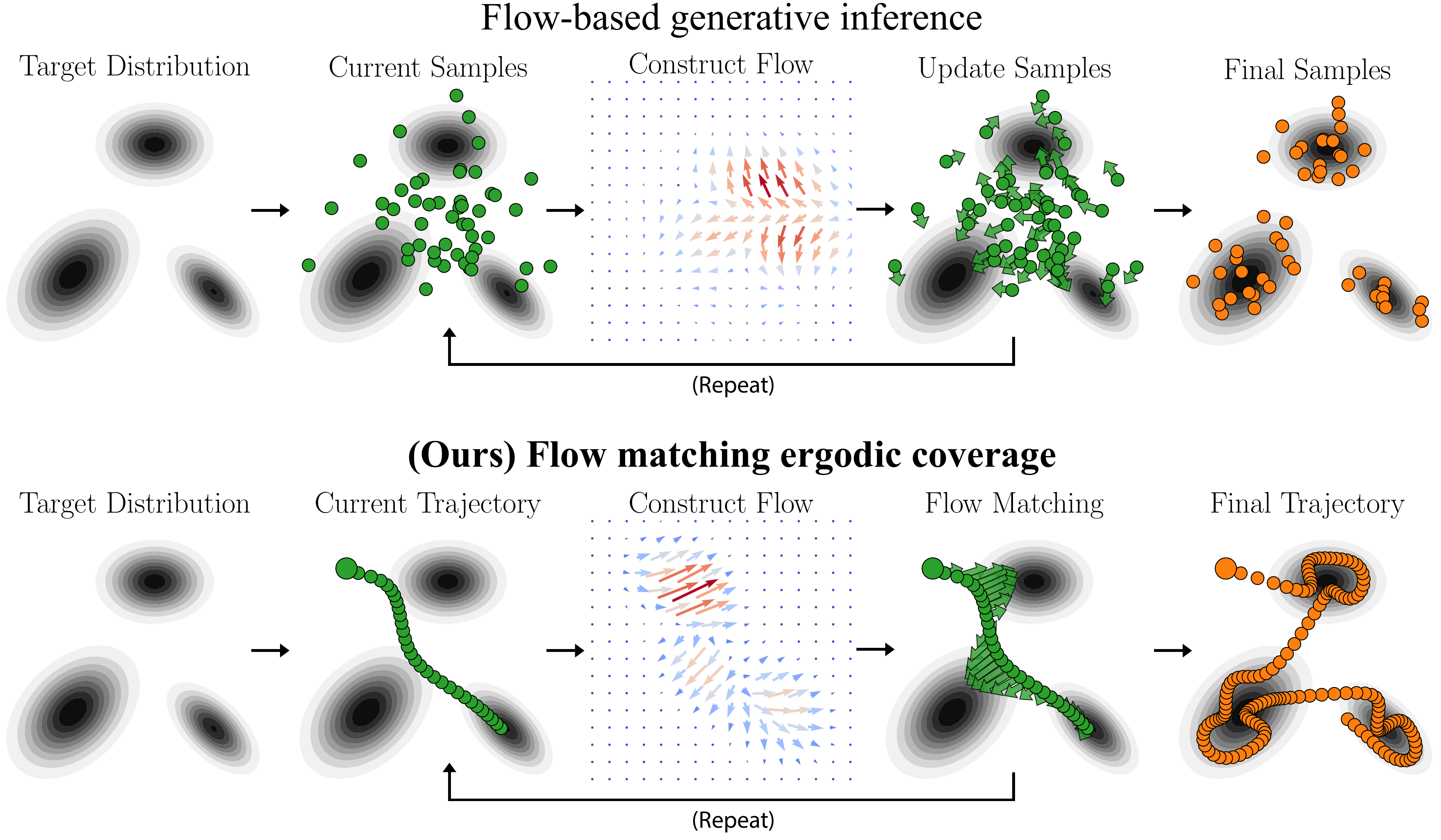}
    \caption{Similarity between flow-based generative inference and flow matching ergodic coverage.}
    \label{fig:overview}
    \vspace{-1em}
\end{figure*}

\section{Introduction}

Many robotic tasks, such as search and rescue~\cite{murphy_human-robot_2004}, wildlife surveys~\cite{shah_multidrone_2020}, and precision agriculture~\cite{norby_path_2024}, require robots to generate motions to explore spatial distributions like thermal signals, population distributions, or soil contaminant levels. Exploration over distributions is challenging, as robots must prioritize high-density regions while also investigating lower-density areas. To address the long-horizon and non-myopic nature of exploration tasks, ergodic coverage generates optimized trajectories whose spatial distribution aligns with the target distribution, providing a formal, information-theoretic notion of coverage that is crucial for exploration tasks. Over the past decade, ergodic coverage has been applied to a wide range of applications, including environmental monitoring~\cite{lanca_model_2024,prabhakar_ergodic_2020}, tactile sensing~\cite{abraham_ergodic_2017,abraham_data-driven_2018}, learning from demonstrations~\cite{kalinowska_ergodic_2021,shetty_ergodic_2022,sun_fast_2025}, shared control~\cite{fitzsimons_ergodic_2022,fitzsimons_task-based_2020}, and embodied active learning~\cite{pinosky_embodied_2024,prabhakar_mechanical_2022}. 

Ergodic coverage solves an optimization problem that minimizes the \emph{ergodic metric}, which measures the difference between the empirical spatial distribution of the trajectory and the target distribution~\cite{miller_ergodic_2016}, while respecting the agent's dynamic constraints. Existing methods are often based on standard trajectory optimization techniques, such as projection-based methods~\cite{miller_trajectory_2013} and augmented Lagrange multiplier methods~\cite{dong_time-optimal_2024}. However, to be compatible with standard trajectory optimization methods, the ergodic metric, even though it evaluates the difference between distributions, must be defined over trajectories. It must also be explicitly formulated and differentiated as runtime cost function. As a result, existing ergodic metrics, such as the most commonly used Fourier ergodic metric~\cite{mathew_metrics_2011} and recently developed kernel-based metrics~\cite{hughes_ergodic_2024,sun_fast_2025}, need to be derived specifically to meet this requirement. On the other hand, commonly used statistical measures in machine learning, such as Kullback-Leibler divergence and optimal transport metrics, are infeasible as the ergodic metric for control synthesis using the standard trajectory optimization methods. This limitation fundamentally restricts the performance of existing ergodic coverage methods.

From a different perspective, ergodic coverage can also be viewed as a sampling process: generating ergodic trajectories resembles sampling from the target distribution, where the samples are constrained spatially and temporally under the embodied agent's dynamics. Recently, continuous flow-based sampling approaches, such as score function-based sampling~\cite{liu_stein_2016,song_score-based_2021} and continuous normalizing flows~\cite{chen_neural_2018}, have gained significant attention due to their accuracy and computational efficiency, as well as their compatibility with a wide range of statistical discrepancy measures, such as the Kullback-Leibler (KL) divergence~\cite{liu_stein_2016} and optimal transport metrics~\cite{genevay_learning_2018}. Furthermore, the flow matching method from~\citet{lipman_flow_2022} enables scalable and efficient optimization of a parameterized flow model based on a reference flow, achieving state-of-the-art performance in generative modeling~\cite{pooladian_multisample_2023,tong_improving_2023}. 

In this work, we bring the advantages of flow-based sampling and flow matching to ergodic coverage to overcome the limitations of existing ergodic metrics and trajectory optimization methods. Our main contributions are twofold: 
\begin{enumerate}
\item We formally derive the flow matching problem from~\citet{lipman_flow_2022} for ergodic coverage and show that it is equivalent to a linear quadratic regulator (LQR) problem. This theoretical result leads to a closed-form iterative optimization algorithm for generating ergodic trajectories with a reference flow from existing flow-based sampling methods (see Fig.~\ref{fig:overview}). 
\item We derive three reference flows for ergodic coverage: the standard Fourier ergodic metric~\cite{mathew_metrics_2011}, the Stein variational gradient flow~\cite{liu_stein_2016}, and the Sinkhorn divergence flow~\cite{genevay_learning_2018}. We show that our method is compatible with the standard ergodic metric and enables alternative ergodic metrics previously infeasible for control synthesis without computational overhead. Compared to state-of-the-art methods, the alternative metrics improve the performance of ergodic coverage over unnormalized distributions and distributions with irregular supports. 
\end{enumerate}
Lastly, we demonstrate our method on a Franka robot through a series of drawing and erasing tasks similar to~\cite{low_drozbot_2022,bilaloglu_tactile_2025}. The code of our implementations and videos of the experiments can be found at our project website: \url{https://murpheylab.github.io/lqr-flow-matching/}. 


\section{Preliminaries and Related Work} \label{sec:prel}

\subsection{Notation}

We denote an $n$-dimensional search space as $\mathcal{X}\subseteq\mathbb{R}^n$. The trajectory of the agent is defined as a mapping $s:[0,T]\mapsto\mathcal{X}$, where $T$ is the time horizon of the trajectory. The trajectory is governed by the agent's dynamics, $\dot{s}(t) = f(s(t), u(t))$, with $u(t)\in\mathcal{U}\subset\mathbb{R}^m$ as the control. The probability density function of the \emph{target distribution} is denoted as $q:\mathcal{X}\mapsto\mathbb{R}_0^+$. We denote the space of all probability distributions over the domain $\mathcal{X}$ as $\mathcal{P}$.

\subsection{Overiew of ergodic coverage}

The goal of ergodic coverage is to match the spatial distribution of the robot trajectory $s(t)$ with the target distribution $q(x)$. The spatial distribution of a trajectory is defined by extending the definition of \emph{empirical distribution}.

\begin{definition}[Trajectory empirical distribution]
    Given a trajectory $s(t)$, the associated empirical distribution is:
    \begin{align}
        p_{[s]}(x) = \frac{1}{T} \int_{0}^{T} \delta(x - s(t)) dt,  \label{eq:emp_distr}
    \end{align} where $\delta(x)$ is a Dirac delta function with the following inner product property:
    \begin{gather}
        \int_{\mathcal{X}} \delta(x{-}s) f(x) dx = f(s), \text{ } \forall s \in \mathcal{X}, \text{ } \forall f\in C(\mathcal{X}),  \label{eq:delta_inner_product}
    \end{gather} the set $C(\mathcal{X})$ contains all continuous functions over $\mathcal{X}$.
\end{definition}

Note that the Dirac delta function $\delta(x)$ is often \emph{heuristically} represented as a point-wise mapping:
\begin{align}
    \delta(x) = \begin{cases}
        +\infty, \text{  if } x=0, \\
        0, \text{ otherwise.}
    \end{cases}
\end{align} But the Dirac delta function is formally a generalized function (also called a distribution) that is not defined as a point-wise mapping but based on the inner product property shown in (\ref{eq:delta_inner_product})~\cite{lighthill_introduction_1958}.

\begin{definition}[Ergodic system \cite{petersen_ergodic_1989,walters_introduction_2000}] \label{eq:ergodic_system_def}
    A dynamic system $s(t)$ is ergodic with respect to $q(x)$ if and only if:
    \begin{align}
        \lim_{T\rightarrow\infty} \int_{\mathcal{X}} g(x) dp_{[s]}(x) = \int_{\mathcal{X}} g(x) dq(x), \text{ } \forall g\in\mathcal{C}(\mathcal{X}).
    \end{align}
\end{definition} 

\begin{lemma} \label{lemma:ergodic_necessary_cond}
    A necessary condition for Definition \ref{eq:ergodic_system_def} is:
    \begin{align}
        \lim_{T\rightarrow\infty} p_{[s]}(x) = q(x), \text{ } \forall x\in\mathcal{X}.
    \end{align}
\end{lemma}

\begin{definition}[Discrepancy measure] \label{def:discrepancy}
    A discrepancy measure $D:\mathcal{P}\times\mathcal{P}\mapsto\mathbb{R}_0^+$ is a non-negative functional, such that $D(p(x),q(x))=0$ if and only if $p(x)=q(x), \forall x\in\mathcal{X}$.
\end{definition}

\begin{definition}[Ergodic metric]
    Given a discrepancy measure $D$, the ergodic metric $E:\mathcal{S}_{T}\times\mathcal{P}\mapsto\mathbb{R}_0^+$ is defined as:
    \begin{align}
        E(s(t),q(x)) = D(p_{[s]}(x), q(x)). \label{eq:og_ergodic_metric}
    \end{align}
\end{definition}

\begin{lemma}
    Based on Lemma~\ref{lemma:ergodic_necessary_cond}, a necessary condition for Definition~\ref{eq:ergodic_system_def} is:
    \begin{align}
        \lim_{T\rightarrow\infty} E(s(t), q(x)) = 0.
    \end{align}
\end{lemma}

\noindent Note that an exact ergodic system exists only at the limit of \emph{infinite} time horizon. In practice, however, ergodic coverage approximately synthesizes an ergodic system with \emph{finite} time horizon $T$ through the following optimization problem.

\begin{definition}[Ergodic coverage]
    \begin{gather}
        u(t)^* = \argmin_{u(t)} E(s(t), q(x)), \label{eq:og_ergodic_control} \\
        \text{s.t. } s(t) = s_0 + \textstyle \int_0^t f(s(t^\prime), u(t^\prime)) dt^\prime, \text{ } t{\in}[0,T]. \nonumber
    \end{gather}
\end{definition}

\noindent Solving (\ref{eq:og_ergodic_control}) is not easy for two main reasons. First, many commonly used discrepancy measures, such as the KL-divergence, are computationally intractable to evaluate and optimize directly. This issue is further complicated by the fact that $p_{[s]}$ is an empirical distribution---where the Dirac delta function cannot be evaluated numerically---and the target distribution $q(x)$ in practice is often represented not as a standard probability density function, but instead in other forms such as unnormalized utility functions, samples, or discrete density grids. Second, optimization of the ergodic metric must respect the dynamics constraints of the system, and system dynamics can be highly nonlinear in practice, which introduces extra computational challenges. 

In the next section, we will review existing ergodic metrics and the corresponding control synthesis methods, focusing on how they address these challenges and their limitations. 

\subsection{Review of existing ergodic coverage methods}

In~\cite{mathew_metrics_2011}, the ergodic metric is specified as a Sobolev space distance using Fourier basis functions, which we name the \emph{Fourier ergodic metric}. This formula offers a computationally tractable approximation of the ergodic metric for numerical optimization by truncating a finite number of Fourier basis functions. A closed-form model predictive control formula with an infinitesimally small planning horizon is proposed to approximate a solution to the constrained trajectory optimization problem. However, this method does not generate optimal ergodic coverage trajectories with a fixed time horizon, which is crucial for practical applications~\cite{dong_time-optimal_2024,sun_fast_2025}.

Alternative trajectory optimization methods have been used to optimize the Fourier ergodic metric over long horizons. A projection-based trajectory optimization algorithm is introduced by~\citet{miller_trajectory_2013-1}, and a method based on augmented Lagrange multiplier, which can incorporate other constraints such as time-optimality and collision avoidance, is proposed by~\citet{dong_time-optimal_2024}. In~\cite{mavrommati_real-time_2018}, a hybrid system-based model predictive control algorithm is introduced with longer planning horizons, which can be extended for multi-agent systems~\cite{abraham_decentralized_2018}. In~\cite{shetty_ergodic_2022}, the evaluation of the Fourier ergodic metric is further accelerated through tensor-train decomposition, especially for applications in high-dimensional spaces. Lastly, it is worth noting that ergodic coverage does not have a unique optimal solution, as there exist multiple or even an infinite number of optimal trajectories with a given ergodic metric. To address this unique property of ergodic coverage, a variational inference-based trajectory optimization framework is proposed in~\citet{lee_stein_2024} using the Stein variational gradient descent algorithm. 

On the other hand, several alternative metrics other than the Fourier ergodic metric have been proposed. A kernelized ergodic metric is proposed by~\cite{sun_fast_2025}, which approximates the $L^2$ distance between distributions under mild regularity conditions and can be optimized through a projection-based algorithm similar to~\cite{miller_trajectory_2013-1}. The kernel ergodic metric has a better scalability compared to the Fourier ergodic metric, and it can be extended to Lie groups. In~\cite{hughes_ergodic_2024-1}, maximum mean discrepancy (MMD), a statistical measure formulated in the reproducing kernel Hilbert space for two-sample tests, is proposed as the ergodic metric. The MMD metric only requires samples from the target distribution instead of probability density functions and can also be extended to Lie groups. Trajectory optimization for the MMD-based metric can be solved using the same augmented Lagrange multiplier-based method in~\cite{dong_time-optimal_2024}. Lastly, an alternative metric is introduced in~\cite{ivic_ergodicity-based_2017} using heat equation, which leads to a better balance between global and local exploration compared to the Fourier ergodic metric, and has been applied to multi-robot aerial survey~\cite{lanca_model_2024} and tactile sensor-based coverage~\cite{bilaloglu_tactile_2025}. However, the trajectory optimization problem is solved through a similar model predictive control formula with an infinitesimally small time horizon as in~\cite{mathew_metrics_2011}. Thus, the resulting trajectories do not generate optimal ergodic coverage performance with a fixed time horizon.

\subsection{Flow-based sampling}

\noindent\textbf{[Overview] } Consider the problem of sampling from a target distribution $q(x)$, while only having access to samples $\{x_i\}$ from an initial distribution $p_0(x)$. One intuition might be to find the optimal transformation $\phi^*(x):\mathcal{X}\mapsto\mathcal{X}$, such that the transformed samples $\{\phi^*(x_i)\}$ match the statistics of the target distribution $q(x)$. 

Furthermore, the transformation $\phi(x)$ can be constructed as a time-dependent transformation $\phi(\tau,x)$ through a time-dependent vector field $g(\tau,x)$ as:
\begin{align}
    \frac{d}{d\tau} \phi(\tau,x) = g(\tau,\phi(\tau,x)), \text{ } \phi_{0}(x) = x, \text{ } \tau\in[0,\mathcal{T}],
\end{align} where $\phi(\tau,x)$ is called a \emph{flow} and $g(\tau,x)$ is the \emph{flow vector field}. The flow creates a \emph{probability density path} $p(\tau,x)$ governed by the Fokker–Planck equation~\cite{jordan_variational_1998}:
\begin{align}
    \frac{d}{d\tau} p(\tau,x) = \nabla \cdot (p(\tau,x) g(\tau,x)), \text{ } p(0,x) = p_0(x) \label{eq:fp_eq}
\end{align} Given the set of samples $\{x_i^\prime\}$ from the initial distribution $p_0(x)$, the flow also creates a path $x_i(\tau)$ for each sample:
\begin{align}
    \frac{d}{d\tau} x_i(\tau) = g(\tau,x_i(\tau)), \text{ } x_i(\tau) = x_i^\prime, \text{ } \tau\in[0,\mathcal{T}]. \label{eq:flow_sample_path}
\end{align} Therefore, flow-based sampling can be formulated as finding the optimal flow vector field to transform the initial distribution $p_0$ toward the target distribution $q(x)$. 

\begin{definition}[Flow-based sampling]
    \begin{align}
        g^*(\tau,x) = \argmin_{g(\tau,x)} \lim_{\tau\rightarrow\mathcal{T}} D(p(\tau), q), \label{eq:flow_sampling_obj} 
    \end{align} where $D$ is a discrepancy measure between distributions.
\end{definition}

\noindent\textbf{[Related work] } Instead of directly solving (\ref{eq:flow_sampling_obj}), most existing flow-based sampling methods construct the flow vector field as a descent direction of the discrepancy measure $D(p,q)$, in which case the transformation of the distribution is equivalent gradient descent in the space of probabilistic measures. The Stein variational gradient descent (SVGD) algorithm~\cite{liu_stein_2017,liu_stein_2016} constructs the Stein variational gradient flow as the steepest descent direction of the KL-divergence in a reproducing kernel Hilbert space. Another widely used family of flow-based sampling methods is based on Wasserstein gradient flows, which originate from the seminal work of~\citet{jordan_variational_1998}. Wasserstein gradient flows construct the steepest descent direction using the Wasserstein metric, where the discrepancy measure $D(p,q)$ can be specified as optimal transport (OT)~\cite{villani_optimal_2009} metrics, maximum mean discrepancy (MMD)~\cite{arbel_maximum_2019}, and more recently the Sinkhorn divergence~\cite{genevay_learning_2018}, which interpolates between MMD and OT metrics~\cite{feydy_interpolating_2019}. 

\subsection{Flow matching}

The flow vector field can also be modeled as a parametric function approximator (e.g., neural networks), which leads to the continuous normalizing flows (CNFs) framework~\cite{chen_neural_2018}. However, the standard maximum likelihood training of CNFs requires numerical simulation of the flow dynamics, which can be prohibitively expensive. To address this issue, \emph{flow matching} is proposed by~\citet{lipman_flow_2022} as a computationally efficient optimization paradigm for CNFs. 

Given a dataset $\{\bar{x}_i\}$ as the samples from the target distribution $q(x)$, the overall goal of CNFs is to learn to generate novel samples from $q(x)$ by transforming samples $\{x_i\}$ from an initial distribution $p_0(x)$, often specified as a Gaussian distribution. The flow matching framework optimizes a parameterized flow vector field $g(\tau,x;\theta)$ through the following optimization problem (equation 5,~\cite{lipman_flow_2022}).
\begin{definition}[Flow matching]
    \begin{align}
        \theta^* = \argmin_{\theta} \mathbb{E}_{\tau, p(\tau,x)} \Vert h(\tau,x) - g(\tau,x;\theta) \Vert^2, \label{eq:flow_matching_obj}
    \end{align} where $p(\tau,x)$ is a probability density path toward the target distribution $q(x)$ under the reference flow vector field $h(\tau,x)$.  
\end{definition}
The reference flow can be constructed using methods from flow-based sampling, with examples including the probability flow ODE based on the score function~\cite{song_score-based_2021} or optimal transport metrics~\cite{tong_improving_2023}. 

The flexibility of flow-based sample generation combined with the computation efficiency of flow matching has emerged as a powerful framework for generative inference, with applications in image generation~\cite{pooladian_multisample_2023,tong_improving_2023}, protein structure generation~\cite{huguet_sequence-augmented_2024,jing_alphafold_2023}, computer vision~\cite{yang_pointflow_2019}, trajectory prediction~\cite{ye_efficient_2024}, and robot policy learning~\cite{chisari_learning_2024}. 

\section{Flow Matching Ergodic Coverage}

\subsection{Problem formulation}

Given a dynamic system $f(s(t), u(t))$ with initial condition $s_0$, we define a \emph{control sequence path} $u(\tau,t)$, in which there are two temporal variables: the \emph{system time} $t{\in}[0,T]$ is the associated with dynamic system (e.g., the time variable for simulating the system trajectory $s(t)$), and the \emph{flow time} $\tau{\in}[0,\mathcal{T}]$ is the time associated with the evolution of the entire control sequence (e.g., the time variable in flow-based sampling). The flow dynamics of the control sequence path $u(\tau,t)$ at any system time $t$ is defined as:
\begin{align}
    \frac{d}{d\tau} u(\tau,t) = v(\tau,t). \label{eq:control_flow}
\end{align} We name $v(\tau,t)$ the \emph{control sequence flow} as it describes the evolution of the control at any system time $t$ across the flow time $\tau$. The control sequence path $u(\tau,t)$ generates a \emph{system trajectory path} $s(\tau,t)$, which is defined at any system time $t$ as:
\begin{align}
    s(\tau,t) {=} s_0 {+} \int_0^t f(s(\tau,t^\prime), u(\tau,t^\prime)) dt^\prime, \text{ } t{\in}[0,T], \tau{\in}[0,\mathcal{T}]. \label{eq:traj_path}
\end{align} We further define the \emph{empirical distribution path} $p_{[s]}(\tau,x)$ based on the system trajectory path (\ref{eq:traj_path}):
\begin{align}
    p_{[s]}(\tau,x) = \frac{1}{T} \int_0^T \delta(x - s(\tau,t)) dt. \label{eq:emp_prob_path}
\end{align} Similar to (\ref{eq:fp_eq}), the flow dynamics of $p_{[s]}(\tau,x)$ is governed by:
\begin{align}
    \frac{d}{d\tau} p_{[s]}(\tau,x) = \nabla \cdot ( p_{[s]}(\tau,x) z(\tau,x)), \label{eq:emp_prob_flow}
\end{align} where $z(\tau,x)$ is the \emph{empirical distribution flow}. It is the flow vector field of the empirical distribution path $p_{[s]}(\tau,x)$, induced by the control sequence flow $v(\tau,t)$ under the dynamics constraints of (\ref{eq:traj_path}). We now define \emph{flow-based ergodic coverage} based on flow-based sampling (\ref{eq:flow_sampling_obj}). 

\begin{definition}[Flow-based ergodic coverage]
    \begin{gather}
        v^*(\tau,t) = \argmin_{v(\tau,t)} \lim_{\tau\rightarrow\mathcal{T}} D(p_{[s]}(\tau,x), q(x)), \label{eq:flow_erogdic_coverage} \\
        \text{s.t. } s(\tau,t) {=} s_0 {+} {\int_0^t} f(s(\tau,t^\prime), u(\tau,t^\prime)) dt^\prime, \text{ } \frac{d}{d\tau} u(\tau,t) {=} v(\tau,t). \label{eq:dyn_constr}
    \end{gather}
\end{definition}

\begin{lemma}
    The solution of flow-based ergodic coverage (\ref{eq:flow_erogdic_coverage}) at $\tau{=}\mathcal{T}$ is the solution of standard ergodic coverage (\ref{eq:og_ergodic_metric}).
\end{lemma}

However, because of the dynamics constraints (\ref{eq:dyn_constr}) between the system trajectory path $s(\tau,t)$ and the control sequence path $u(\tau,t)$, the above optimization problem cannot be solved directly. Instead, we formulate a \emph{flow matching} problem for ergodic coverage based on (\ref{eq:flow_matching_obj}).

\begin{definition}[Flow matching ergodic coverage] 
    \begin{gather}
        v^*(\tau,t) = \argmin_{v(\tau,t)} \mathbb{E}_{p_{[s]}(\tau,x)} \Vert h(\tau,x) - z(\tau,x) \Vert^2, \text{ } \forall \tau{\in}[0{,}\mathcal{T}], \label{eq:erg_flow_matching_obj}
    \end{gather} where $h(\tau,x)$ is a reference flow vector field that generates a probability density path toward $q(x)$ from $p_{[s]}(\tau,x)$.  
\end{definition} The reference flow can be constructed the same way as in existing flow-based sampling methods and we will specify three reference flows in Section~\ref{subsec:flow_spec}. 

The intuition behind the flow matching ergodic coverage formulation (\ref{eq:erg_flow_matching_obj}) is to find the optimal \emph{control sequence flow} $v(\tau,t)$ such that the induced empirical distribution flow $z(\tau,x)$ closely match the reference flow $h(\tau,x)$ at any flow time $\tau$, thus generating a path of the trajectory empirical distribution toward the target distribution. However, to solve the flow matching problem, we still need to derive the relationship between the control sequence flow $v(\tau,t)$ and the induced empirical distribution flow $z(\tau,x)$ under the dynamics constraints (\ref{eq:dyn_constr}). In the next section, we will show that $z(\tau,x)$ and $v(\tau,x)$ are governed by a linear dynamic system. Therefore, (\ref{eq:erg_flow_matching_obj}) is equivalent to a linear quadratic regulator (LQR) problem---which we name \emph{linear quadratic flow matching}---that can be solved in closed-form at any flow time $\tau$.

\begin{algorithm} [t!]
    \caption{Flow matching ergodic coverage}
    \label{algo:ergodic_coverage}
    \begin{algorithmic}[1] 
        \Procedure{FlowMatching}{$s_0$, $u_0(t)$, $\eta$, $\Delta\tau$}
        \State $\tau \gets 0$
        \While{not converged}
            \State Simulate $s(\tau,t)$ from $s_0$ and $u(\tau,t)$ with $\tau$ fixed
            \State Evaluate the reference flow $h(\tau,s(\tau,t))$ at $\tau$
            \State Solve (\ref{eq:lq_flow_matching}) for $v(\tau,t)$
            \State $u_{\tau{+}\Delta\tau}(t) \gets u(\tau,t) + \eta \cdot v(\tau,t)$
            \State $\tau\gets\tau+\Delta\tau$
        \EndWhile
        \State \textbf{return} $s(\tau,t)$ and $u(\tau,t)$
        \EndProcedure
    \end{algorithmic}
\end{algorithm}

\subsection{Linear quadratic flow matching}

We start the derivation of the linear quadratic flow matching problem by substituting the definition of $p_{[s]}(\tau,x)$ into the flow matching objective (\ref{eq:erg_flow_matching_obj}):
\begin{align}
    & \mathbb{E}_{p_{[s]}(\tau,x)} \Vert h(\tau,x) - z(\tau,x) \Vert^2 \nonumber \\
    & = \int_{\mathcal{X}} \left( \frac{1}{T} \int_0^T \delta(x {-} s(\tau,t)) dt \right) \Vert h(\tau,x) - z(\tau,x) \Vert^2  dx \nonumber \\
    & = \frac{1}{T} \int_0^T \left( \int_{\mathcal{X}} \delta(x {-} s(\tau,t)) \Vert h(\tau,x) - z(\tau,x) \Vert^2 dx \right) dt \nonumber \\
    & = \frac{1}{T} \int_0^T \Vert h(\tau,s(t)) - z(\tau,s(t)) \Vert^2  dt \nonumber \\
    & = \frac{1}{T} \int_0^T \Vert \Tilde{h}(\tau,t) - \Tilde{z}(\tau,t) \Vert^2  dt.
\end{align} 

We now derive the system time dynamics of $\Tilde{z}(\tau,t)$ under the control sequence flow $v(\tau,t)$ at any flow time $\tau$. As defined in (\ref{eq:emp_prob_flow}), $z(\tau,x)$ is the flow vector field of $p_{[s]}(\tau,x)$, since the probability density path $p_{[s]}(\tau,x)$ is an empirical distribution path (\ref{eq:emp_prob_path}) over the system trajectory path $s(\tau,t)$ (\ref{eq:traj_path}), we have the following based on (\ref{eq:flow_sample_path}):
\begin{align}
    \frac{d}{d\tau} s(\tau,t) = z(\tau,s(\tau,t)) = \Tilde{z}(\tau,t).
\end{align} Furthermore, the system trajectory path $s(\tau,t)$ is governed by the dynamics shown in (\ref{eq:traj_path}) and the control sequence path $u(\tau,t)$ is governed by the ODE shown in (\ref{eq:control_flow}). Therefore, with a sufficiently small perturbation $\epsilon{>}0$ at a fixed $\tau$, we have:
\begin{align}
    & s(\tau,t) + \epsilon \cdot \Tilde{z}(\tau,t) \nonumber \\
    & = s_0 {+} \int_0^t f\Big(s(\tau,t^\prime) + \epsilon \cdot \Tilde{z}(\tau,t^\prime), u(\tau,t^\prime) + \epsilon \cdot v(\tau,t^\prime) \Big) dt^\prime, \nonumber
\end{align} from which\footnote{Note that there is no perturbation on the initial state $s_0$ since it is fixed.} we have the dynamics of $\Tilde{z}(\tau,t)$ as:
\begin{align}
    & \Tilde{z}(\tau,t) = \frac{d}{d\epsilon} \left[ s(\tau,t) + \epsilon \cdot \Tilde{z}(\tau,t) \right]_{\epsilon=0} \nonumber \\
    & = \frac{d}{d\epsilon} \left[ \int_0^t f\Big(s(\tau,t^\prime) {+} \epsilon {\cdot} \Tilde{z}(\tau,t^\prime), u(\tau,t^\prime) {+} \epsilon {\cdot} v(\tau,t^\prime) \Big) dt^\prime \right]_{\epsilon=0} \nonumber \\
    & = \int_0^t \frac{d}{d\epsilon} \left[  f\Big(s(\tau,t^\prime) {+} \epsilon {\cdot} \Tilde{z}(\tau,t^\prime), u(\tau,t^\prime) {+} \epsilon {\cdot} v(\tau,t^\prime) \Big) \right]_{\epsilon=0} dt^\prime \nonumber \\
    & = \int_0^t A(\tau,t^\prime) {\cdot} \Tilde{z}(\tau,t^\prime) + B(\tau,t^\prime) {\cdot} v(\tau,t^\prime) dt^\prime, \label{eq:linear_dyn} 
\end{align} where 
\begin{align}
    A(\tau,t^\prime) & = \frac{\partial}{\partial s}f(s(\tau,t^\prime), u(\tau,t^\prime)), \nonumber \\
    B(\tau,t^\prime) & = \frac{\partial}{\partial u}f(s(\tau,t^\prime), u(\tau,t^\prime)). \nonumber
\end{align} From (\ref{eq:linear_dyn}), we can see that the system time dynamics of $\Tilde{z}(\tau,t)$ and $v(\tau,t)$ are governed by a \emph{time-varying linear system} at a fixed flow time $\tau$, which is the linearization of the system dynamics $f(s(t), u(t))$. We can now define the \emph{linear quadratic flow (LQ) matching} problem based on this result. 

\begin{figure*}[htbp] 
    \centering
    \includegraphics[width=0.99\textwidth]{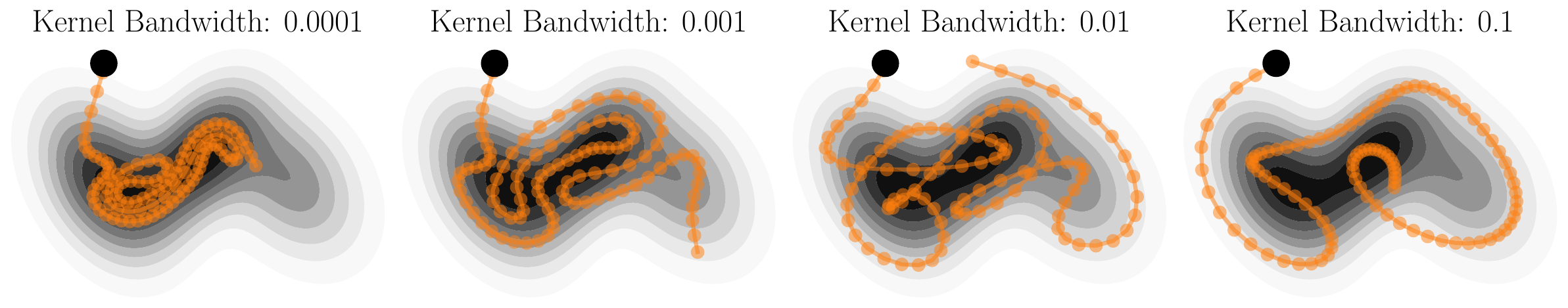}
    \caption{The bandwidth of the RBF kernel in Stein variational gradient flow affects the resulting ergodic trajectories. Overly small bandwidth could cause the mode collapse issue while overly large bandwidth could also lead to under-exploration.}
    \label{fig:svgd_parameters}
\end{figure*}

\begin{definition}[Linear quadratic flow matching] At any flow time $\tau$, the problem is defined as
    \begin{gather}
        v^*(\tau,t) = \argmin_{v(\tau,t)} \int_0^T \Vert \Tilde{h}(\tau,t) {-} \Tilde{z}(\tau,t) \Vert^2_Q + \Vert v(\tau,t) \Vert^2_R dt \label{eq:lq_flow_matching} \\
        \text{s.t. } \Tilde{z}(\tau,t) = \int_0^t A(\tau,t^\prime) {\cdot} \Tilde{z}(\tau,t^\prime) + B(\tau,t^\prime) {\cdot} v(\tau,t^\prime) dt^\prime,
    \end{gather} where $Q$ and $R$ are regularization matrices.
\end{definition}

The LQ flow matching problem (\ref{eq:lq_flow_matching}) follows the standard linear quadratic regulator formula; thus, it is convex and can be solved in closed-form by solving the continuous-time Riccati equation~\cite{hauser_projection_2002}. To generate an ergodic coverage trajectory, the linear quadratic flow matching problem can be solved at a given flow time $\tau$ (starting from $\tau{=}0$) to generate the optimal control sequence flow $v^*(\tau,t)$, from which we can forward simulate the control sequence path $u(\tau,t)$ using Euler's method:
\begin{align}
    u(\tau{+}\Delta\tau, t) = u(\tau,t) + \eta \cdot v^*(\tau,t),
\end{align} where $\Delta\tau$ and $\eta$ are the integration step size for the flow time and the control sequence path, respectively. We summarize the overall process in Algorithm~\ref{algo:ergodic_coverage}. Note that the projection-based trajectory optimization method from~\cite{miller_trajectory_2013-1} also solves an LQR problem in each iteration, but the LQR formulation in~\cite{miller_trajectory_2013-1} is different from ours (\ref{eq:lq_flow_matching}) and must be analytically derived for a specific ergodic metric, while our linear quadratic flow matching formula is compatible with any reference flow without adaptation. Lastly, some flow specifications, such as in continuous normalizing flows, require a finite terminal flow time $\mathcal{T}$, while others, such as Stein variational gradient flows, converge as $\mathcal{T}{\rightarrow}\infty$. The latter case reduces the dependency on explicit integration of the flow time $\tau$, which improves the numerical stability.  In the next section, we will provide three specifications for the reference flow, none of which requires a finite flow time range $\mathcal{T}$.

\begin{figure*}[htbp] 
    \centering
    \includegraphics[width=\textwidth]{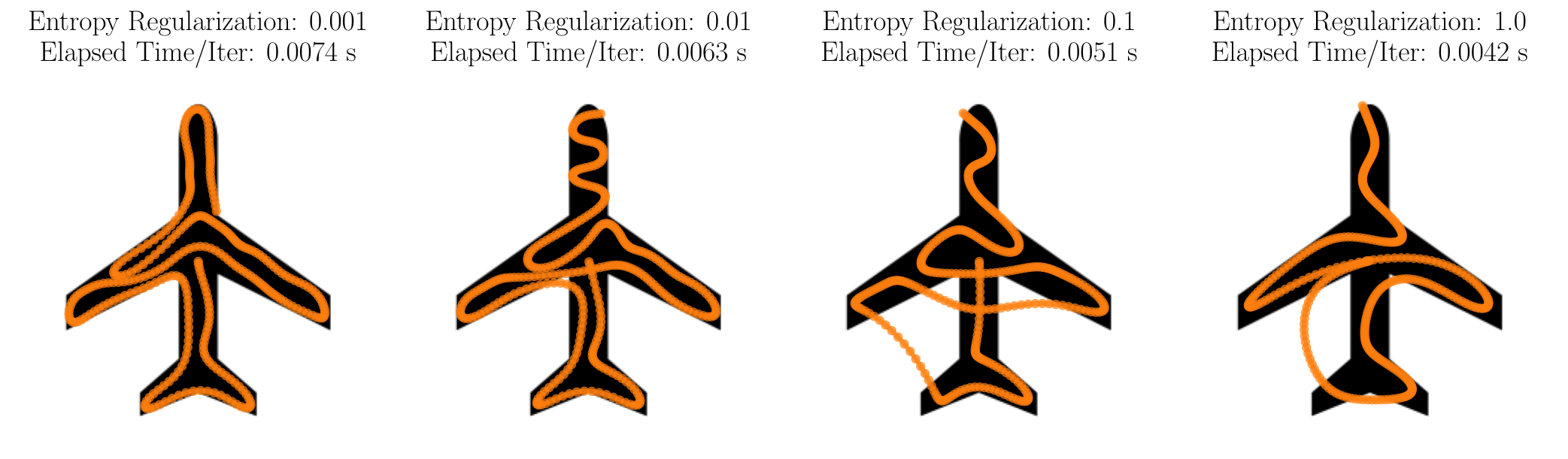}
    \caption{The entropy regularization weight $\epsilon$ in Sinkhorn divergence (\ref{eq:sinkhorn_divergence}) affects the resulting ergodic trajectories. Smaller values make the divergence closer to the Wasserstein distance, improving coverage performance at the cost of slower computation~\cite{feydy_interpolating_2019}.}
    \label{fig:sinkhorn_parameters}
    \vspace{-1em}
\end{figure*}

\subsection{Specifications of reference flows} \label{subsec:flow_spec}

We first derive the reference flow for the widely used Fourier ergodic metric from~\citet{mathew_metrics_2011}. We then specify the reference flow using the Stein variational gradient flow~\cite{liu_stein_2016} and Sinkhorn divergence flow~\cite{feydy_interpolating_2019}.

\noindent\textbf{[Fourier ergodic metric flow] } Without loss of generality, we define the normalized Fourier basis function over a $n$-dimensional rectangular space $\mathcal{S}{=}[0,L_1]{\times\cdots\times}[0,L_n]$:
\begin{align}
    f_{\mathbf{k}}(x) = \frac{1}{h_{\mathbf{k}}} \prod_{i=1}^{n} \cos\left( \frac{k_i\pi}{L_i} x_i \right)
\end{align} where
\begin{align}
        & x = [x_1, x_2, \cdots, x_n] \in \mathcal{S} \nonumber \\
        & \mathbf{k} = [k_1, \cdots, k_n] \in \mathcal{K} \subset \mathbb{N}^n \nonumber \\
        & \mathcal{K} = [0, 1, \cdots, K_1]\times\cdots\times [0, 1, \cdots, K_n], \nonumber
\end{align} and $h_{\mathbf{k}}$ is the normalization term such that the norm of each basis function is $1$. The Fourier ergodic metric is defined based on the Sobolev distance between two distributions:
\begin{gather}
    D(p,q) = \sum_{\mathbf{k}\in\mathcal{K}} \lambda_k (p_{\mathbf{k}} - q_{\mathbf{k}})^2, \\
    p_{\mathbf{k}} {=} \langle p(x), f_{\mathbf{k}}(x) \rangle, \text{ } q_{\mathbf{k}} {=} \langle q(x), f_{\mathbf{k}}(x) \rangle, \text{ } \lambda_{\mathbf{k}} {=} (1 + \Vert\mathbf{k}\Vert)^{-\frac{n+1}{2}}. \nonumber
\end{gather} We can substitute $p(x)$ as the trajectory empirical distribution $p_{[s]}(\tau,x)$, which leads to the following based on the inner product property of the Dirac delta function:
\begin{align}
    p_{\mathbf{k}}(\tau) = \langle p_{[s]}(\tau,x), f_{\mathbf{k}}(x) \rangle = \frac{1}{T} \int_0^T f_{\mathbf{k}} (s(\tau,t)) dt.
\end{align} Following~\cite{ambrosio_gradient_2008}, we can derive the reference flow for the Fourier ergodic metric as a Wasserstein gradient flow:
\begin{align}
    h(\tau,x) & = \frac{d}{dx} \left[ \frac{\partial}{\partial p} D(p_{[s]}(\tau,x),q(x)) \right] \nonumber \\
    & = \frac{d}{dx} \left[ \sum_{\mathbf{k}\in\mathcal{K}} 2\lambda_k (p_{\mathbf{k}}(\tau)-q_{\mathbf{k}}) f_{\mathbf{k}}(x) \right] \nonumber \\
    & = \sum_{\mathbf{k}\in\mathcal{K}} 2\lambda_k (p_{\mathbf{k}}(\tau)-q_{\mathbf{k}}) \cdot \frac{d}{dx} f_{\mathbf{k}}(x).
\end{align} Note that, the reference flow for the Fourier ergodic metric can be evaluated in closed-form.

\noindent\textbf{[Stein variational gradient flow] } The Stein variational gradient flows optimizes the KL-divergence between $p_{[s]}(\tau,x)$ and the target distribution $q(x)$: 
\begin{align}
    D_{KL}(p_{[s]}(\tau,x),q(x)) = \mathbb{E}_{p_{[s]}(\tau,x)}\left[ \log\left( \frac{p_{[s]}(\tau,x)}{q(x)} \right) \right].
\end{align} It is defined as:
\begin{align}
    h(\tau,x) = \mathbb{E}_{p_{[s]}(\tau,x^\prime)} \left[ \mathcal{A}_{q}k(x,x^\prime) \right], \label{eq:stein_flow}
\end{align} where $k(x,x^\prime)$ is a kernel function\footnote{It is often specified as a radial basis function (RBF) in practice.} and $\mathcal{A}$ is the Stein operator~\cite{liu_stein_2016} defined as:
\begin{align}
    \mathcal{A}_{q}k(x, x^\prime) = k(x,x^\prime) \frac{d}{dx} \log q(x) + \frac{d}{dx} k(x, x^\prime). \label{eq:stein_operator}
\end{align} The Stein reference flow only requires access to the derivative of the log-likelihood function of the target distribution---also called the score function of the target distribution. Furthermore, the expectation term in (\ref{eq:stein_flow}) can be calculated as:
\begin{align}
    \mathbb{E}_{p_{[s]}(\tau,x^\prime)} \left[ \mathcal{A}_{q}k(x,x^\prime) \right] = \frac{1}{T} \int_0^T \mathcal{A}_{q}k(x,s(\tau,t)) dt.
\end{align} In~\citet{liu_stein_2016}, it is shown that the Stein variational gradient flow is the steepest descent direction for the KL-divergence between $p_{[s]}(\tau,x)$ and $q(x)$ in a reproducing kernel Hilbert space.

The key advantage of the Stein variational gradient flow is that, it requires evaluation of the score function---the derivative of the log-likelihood of the target distribution. The evaluation of score function does not require the target distribution to have a normalized or even non-negative density function. This property has been shown to significantly improve the numerical stability of generative models in~\citet{song_score-based_2021}. Similarly, this property is also crucial for applications of ergodic coverage as in practice, where the target distributions are often unnormalized utility functions, such as from Gaussian processes regression~\cite{abraham_data-driven_2018,lee_stein_2024}. On the other hand, similar to Stein variational gradient descent SVDG), the choice of kernel function parameters affects the resulting ergodic trajectories. As shown in Fig.~\ref{fig:svgd_parameters}, in the case of a radial basis kernel function, an overly small bandwidth could lead to the mode collapse issue commonly seen in SVGD~\cite{zhuo_message_2018,lee_stein_2024} and an overly large bandwidth could result in trajectories that under explore regions with high density. Instead of hand-tuning the kernel parameters, a parameter selection formula is given in~\citet{liu_stein_2016} for radial basis kernel functions (See Section 5 of~\cite{liu_stein_2016}). 

Lastly, the use of Stein variational gradient flow in our method is to replace the standard Fourier ergodic metric to optimize a single trajectory. At the same time, Stein variational inference is also applied to ergodic coverage to generate multiple optimal trajectories under the Fourier ergodic metric in~\citet{lee_stein_2024}.

\begin{figure*}[htbp] 
    \centering
    \begin{subfigure}[t]{0.37\textwidth}
        \centering
        \includegraphics[width=\linewidth, keepaspectratio]{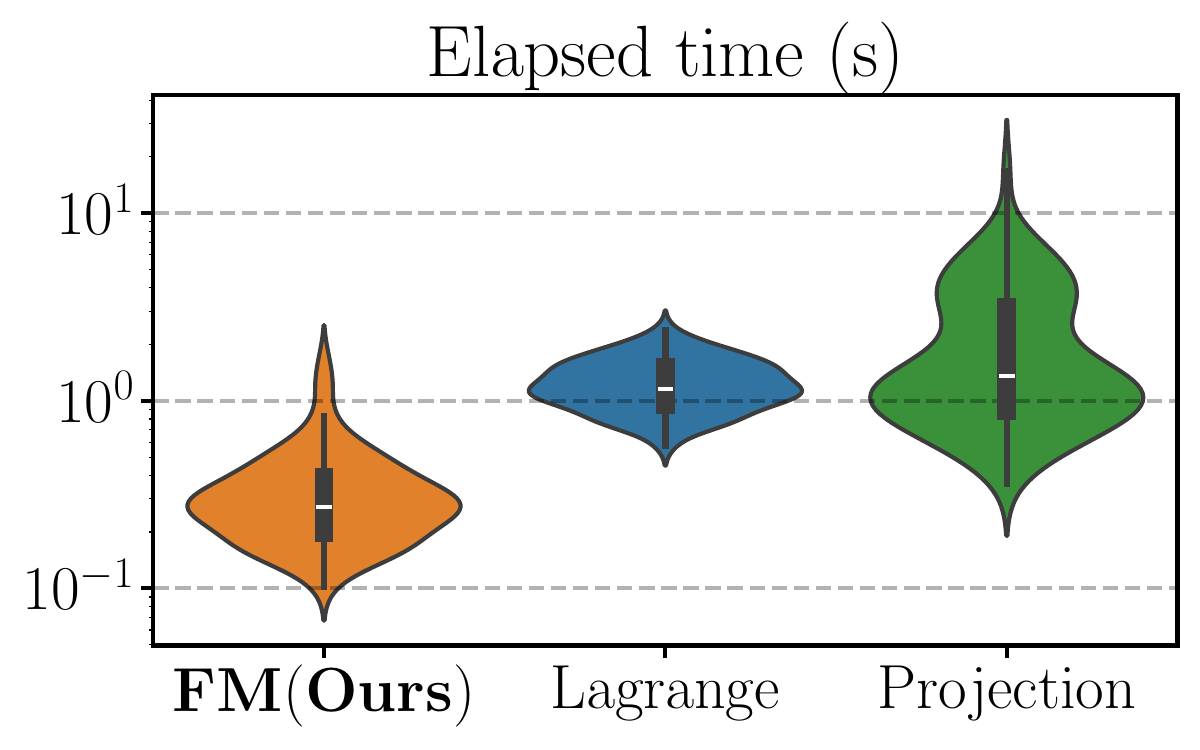} 
    \end{subfigure}
    \hfill
    \begin{subfigure}[t]{0.61\textwidth}
        \centering
        \includegraphics[width=\linewidth, keepaspectratio]{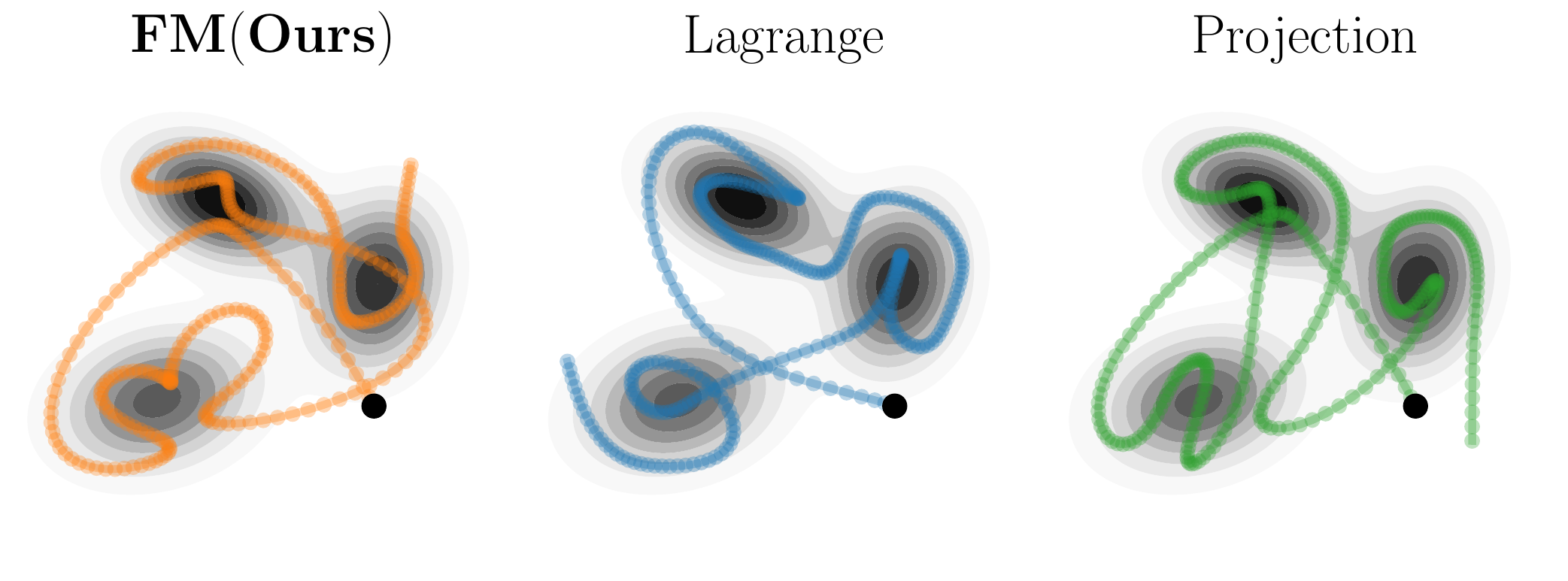} 
    \end{subfigure}
    \caption{\textbf{Results for Benchmark Q1}. Quantitative (left) and qualitative (right) results for the comparison between our method with existing trajectory optimization methods for the Fourier ergodic metric. Our method consistently reaches the same desired level of ergodicity with less time. The white line in the violin plot is the median of the results, and the black dot in the trajectory plot is the initial position.}
    \label{fig:comparison_fourier_trajopt}
    \vspace{-1em}
\end{figure*}

\noindent\textbf{[Sinkhorn divergence flow] } The Sinkhorn divergence is a discrepancy measure based on the entropic regularized optimal transport distance $OT_{\omega}(p,q)$ between two distributions:
\begin{gather}
    OT_{\omega}(p,q) {=} \min_{\pi(x,x^\prime)} \mathbb{E}_{\pi} \Big[ c(x,x^\prime) \Big] + \omega KL(\pi(x,x^\prime)\Vert p(x)q(x^\prime)), \nonumber \\
    \text{s.t. } \int \pi(x,x^\prime) dx^\prime = p(x) \text{ and } \int \pi(x,x^\prime) dx = q(x^\prime), \nonumber
\end{gather} where the transport plan $\pi(x,x^\prime)$ is a joint probability distribution, $\omega$ is a regularization weight term, and $c(x,x^\prime)$ is a cost function\footnote{It is often specified as the $L^p$ distance between $x$ and $x^\prime$ in practice.}. The entropic regularized OT distance suffers from entropic bias---the minima of $OT_{\omega}(p,q)$ is not $0$ with $\omega{>}0$. The Sinkhorn divergence~\cite{genevay_learning_2018} removes the entropic bias:
\begin{align}
    D_{\omega}(p,q) = OT_{\omega}(p,q) - \frac{1}{2} OT_{\omega}(p,p) - \frac{1}{2} OT_{\omega}(q,q). \label{eq:sinkhorn_divergence}
\end{align} 

While evaluating the Sinkhorn divergence directly over continuous probability density functions is computationally intractable, it can be evaluated efficiently over samples from the distributions using the Sinkhorn algorithm~\cite{cuturi_sinkhorn_2013}, which is an iterative algorithm based on matrix scaling. Furthermore, since all the calculations of the Sinkhorn algorithm are differentiable, the gradient of the Sinkhorn divergence can be evaluated over a batch of samples using auto-differentiation~\cite{genevay_learning_2018}. For flow matching ergodic coverage, we can discretize the system trajectory path $s(\tau,t)$ at discrete system time steps $\{t_i\}$. At a given flow time $\tau$ and a set of samples $\{x_i^\prime\}$ from the target distribution $q(x)$, we can directly evaluate the reference flow $h(\tau,x)$ at all the discrete system time steps in a batch:
\begin{align}
    \{ \Tilde{h}(\tau,t_i) \}_i = \textit{AutoDiff}_{\{s(\tau,t_i)\}_i}\left( D_{\omega}\Big(\{s(\tau,t_i)\}_i, \{x_j^\prime\}_j\Big) \right). \nonumber
\end{align} We refer the readers to~\cite{genevay_learning_2018} for more details on the calculation of the Sinkhorn divergence and the Sinkhorn algorithm. In our experiments, we use the Geomloss package from~\cite{feydy_interpolating_2019} to efficiently evaluate the Sinkhorn divergence and the associated flows, which supports accelerated evaluations on GPUs.

As an optimal transport metric, the Wasserstein distance is particularly sensitive to the deformation of the distributions' supports~\cite{feydy_interpolating_2019}, which makes Sinkhorn divergence flow particularly advantageous with non-smooth target distributions with local geometric features, such as distributions with irregular supports. On the other hand, the entropy regularization term $\epsilon$ in (\ref{eq:sinkhorn_divergence}) affects the resulting ergodic trajectories. As shown in Fig.~\ref{fig:sinkhorn_parameters}, smaller values of entropy regularization make Sinkhorn divergence closer to the Wasserstein distance, improving the coverage performance at the cost of slower computation~\cite{feydy_interpolating_2019}.

\section{Numerical Benchmark}

\subsection{Overview}

We aim to answer the following four questions in our numerical benchmarks:

\begin{enumerate}[label=Q\arabic*:]
    \item For the Fourier ergodic metric, how does our flow matching framework perform compared to existing trajectory optimization methods in terms of ergodic coverage results and numerical efficiency?
    \item By integrating Stein variational gradient flows, does our flow matching framework improve the robustness of ergodic coverage when facing inaccurately normalized target distributions?
    \item By integrating Sinkhorn divergence flows, does our flow matching framework improve the ergodic coverage results over non-smooth target distributions?
    \item How does our flow matching framework handle different linear or nonlinear dynamics?
    \item How does the trajectory horizon affect the computation of each reference flow?
\end{enumerate}


\subsection{Baseline selection and implementation details}

For Q1, we compare our method to two Fourier-based ergodic trajectory optimization methods: the projection-based trajectory optimization method from~\citet{miller_trajectory_2013-1} and the augmented Lagrange multiplier-based method from~\cite{dong_time-optimal_2024}. For Q2, we use the Fourier ergodic metric and the kernelized ergodic metric from \citet{sun_fast_2025}, since both metrics and the Stein variational gradient flow require access to the probability density function of the target distribution. For the non-smooth distributions in Q3, we use the Fourier ergodic metric and the maximum mean discrepancy (MMD)-based ergodic metric from~\citet{hughes_ergodic_2024-1}; the Fourier ergodic metric can be calculated over discretized target distribution representations (e.g., images), and the MMD-based metric is designed for sample-based distribution representations.

The computation related to the Fourier ergodic metric is implemented in Python using the JAX package for acceleration. We implement the continuous-time Riccati equation solver for the linear quadratic flow matching problem (\ref{eq:lq_flow_matching}) in C++ with Python interface, same as the projection-based trajectory optimization algorithm. We use the JAX-based implementation from~\citet{dong_time-optimal_2024} for the augmented Lagrange multiplier-based trajectory optimization method. The computation related to Sinkhorn divergence is implemented in Python using the Geomloss package from~\citet{genevay_learning_2018}. The code of our implementation will be released on the project website: \url{https://murpheylab.github.io/lqr-flow-matching/}. 

\begin{figure*}[htbp] 
    \centering
    \begin{subfigure}[t]{0.37\textwidth}
        \centering
        \includegraphics[width=\linewidth, keepaspectratio]{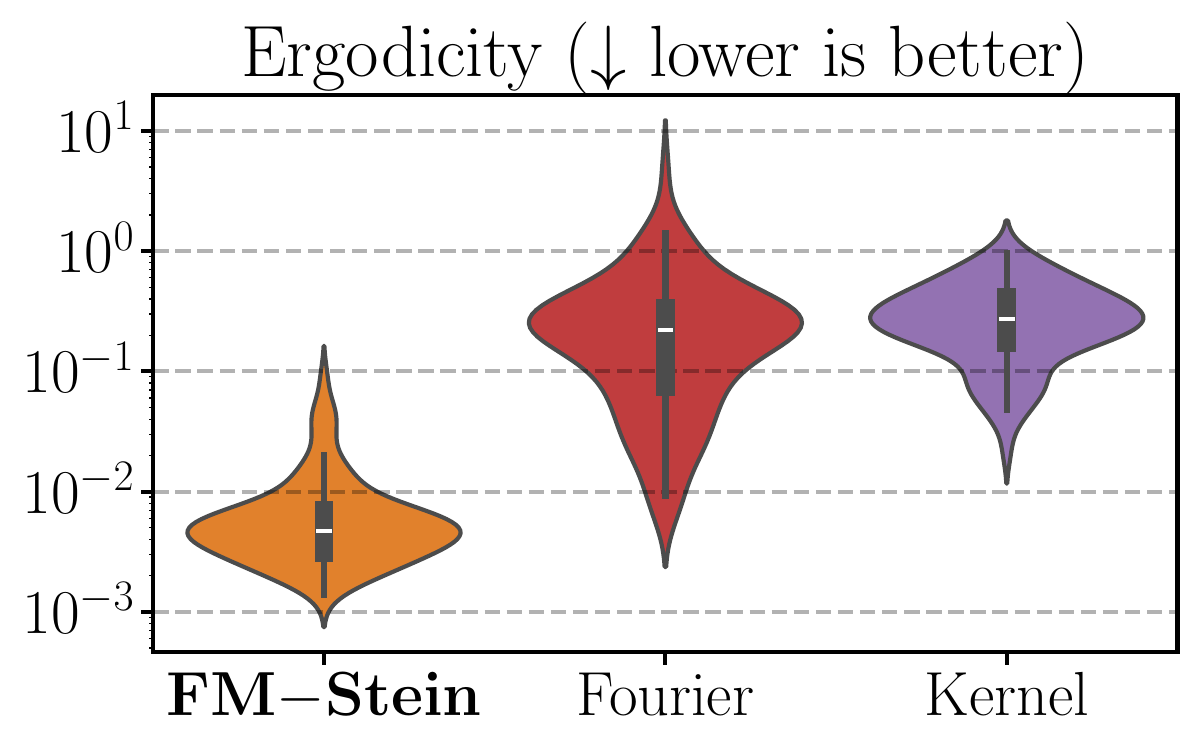} 
    \end{subfigure}
    \hfill
    \begin{subfigure}[t]{0.62\textwidth}
        \centering
        \includegraphics[width=\linewidth, keepaspectratio]{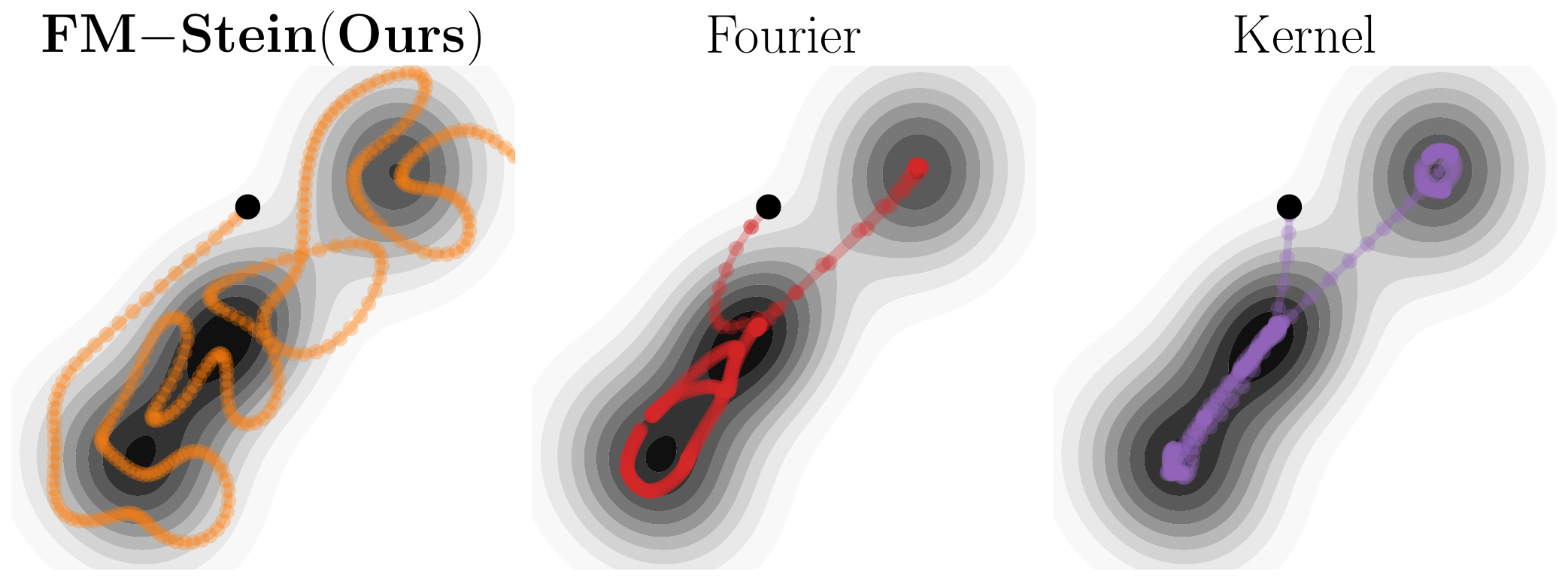} 
    \end{subfigure}
    \caption{\textbf{Benchmark Q2.A}. Quantitative (left) and qualitative (right) results show that our method is more accurate and consistent under normalization errors compared to SOTA methods. The white line in the violin plot is the median of the results, and the black dot in the trajectory plot is the initial position.}
    \label{fig:comparison_fourier_stein}
\end{figure*}

\begin{figure*}[htbp] 
    \centering
    \begin{subfigure}[t]{0.35\textwidth}
        \centering
        \includegraphics[width=\linewidth, keepaspectratio]{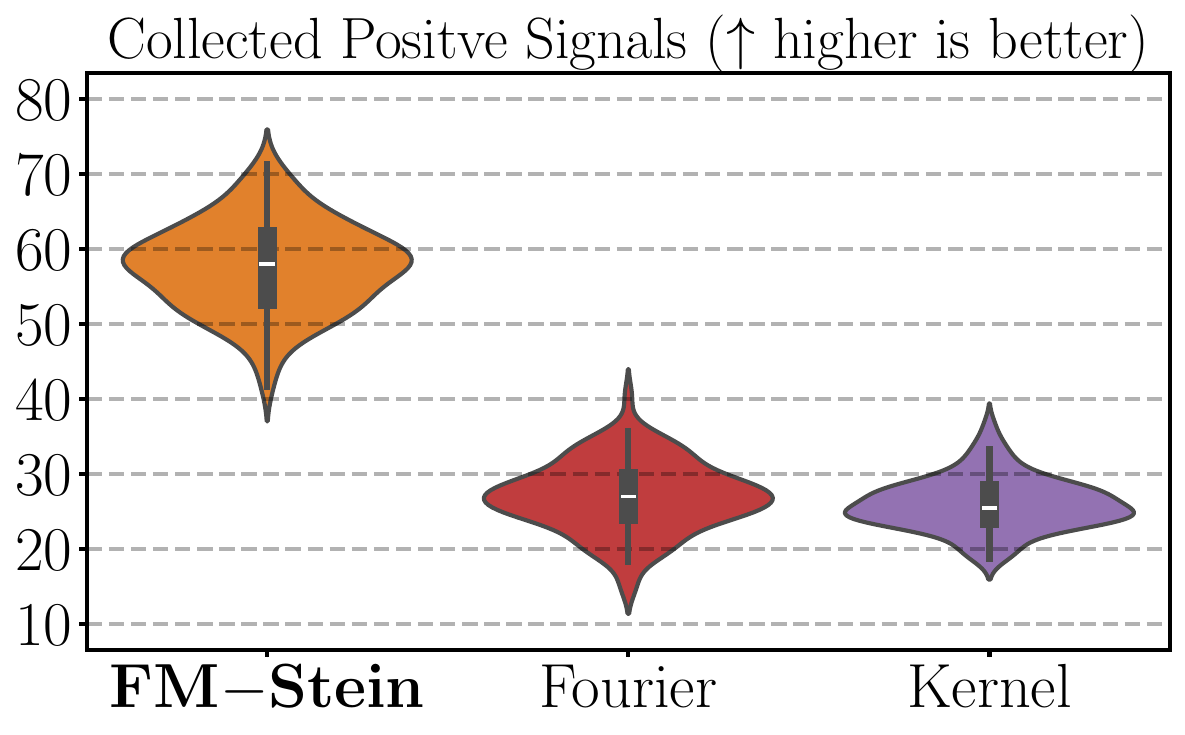} 
    \end{subfigure}
    \hfill
    \begin{subfigure}[t]{0.62\textwidth}
        \centering
        \includegraphics[width=\linewidth, keepaspectratio]{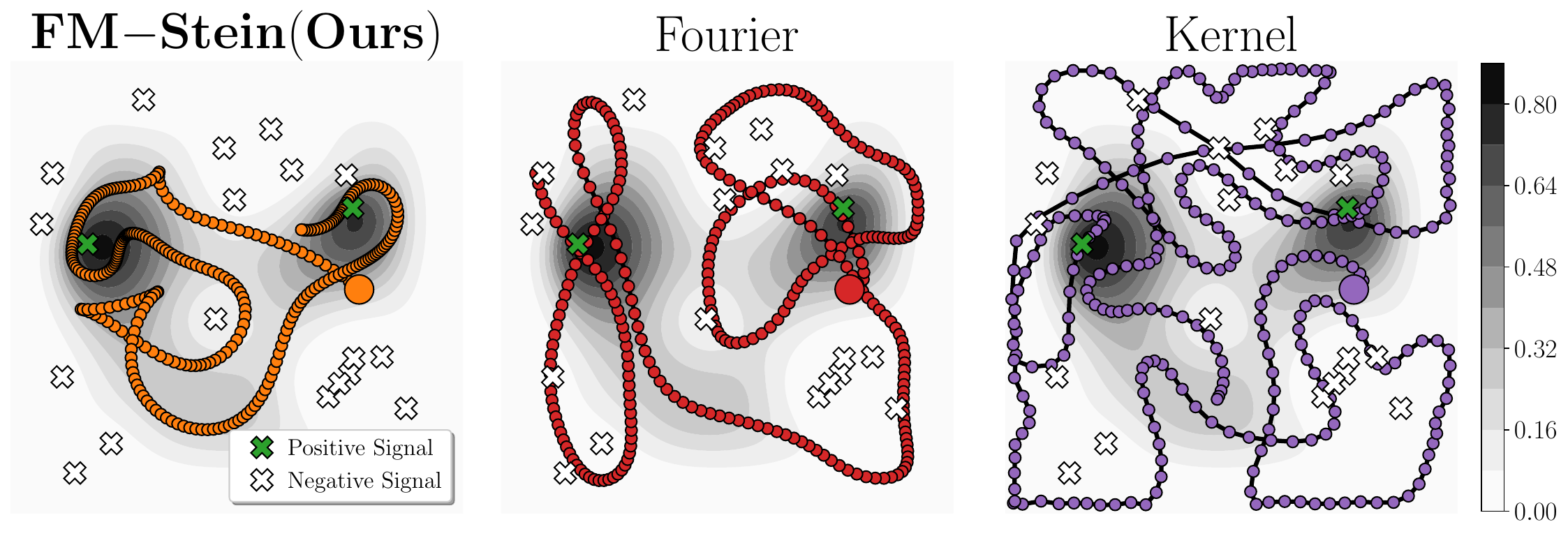} 
    \end{subfigure}
    \caption{\textbf{Benchmark Q2.B}. Quantitative (left) and qualitative (right) results show that our method is consistently collects more positive life signals compared to SOTA methods with unnormalized target distributions. The white line in the violin plot is the median of the results, and the black dot in the trajectory plot is the initial position.}
    \label{fig:comparison_sns}
    \vspace{-1em}
\end{figure*}

\subsection{Experiment results for Q1}

\noindent\textbf{[Benchmark Q1 design] } We randomly generate a trimodal Gaussian mixture distribution and the initial position in each trial, and benchmark our flow matching framework (\textsf{FM}) alongside projection-based trajectory optimization (\textsf{Projection}) and augmented Lagrange multiplier-based trajectory optimization (\textsf{Lagrange}) over 100 trials. We set a desired value of the Fourier ergodic metric $0.005$, and measure the wall clock time required for each method to reach the desired level of ergodicity (in seconds). We use second-order point mass dynamics for all the tests.

\noindent\textbf{[Q1 results] } The statistics of the elapsed times for all the methods across the 100 randomized trials are shown in Fig.~\ref{fig:comparison_fourier_trajopt} (left). The results show that our flow matching framework consistently reaches the desired level of ergodicity in less time. In Fig.~\ref{fig:comparison_fourier_trajopt} (right), we also show qualitative comparisons of the ergodic trajectories generated by each method over the same target distribution and under the same initial condition.

\subsection{Experiment results for Q2}

We answer Q2 with two numerical benchmarks (Q2.A and Q2.B). Details of benchmark design and results are as follow.

\noindent\textbf{[Benchmark Q2.A design] } We follow the same randomization process in Q1. In addition, we randomly scale the target distribution between $0.1$ and $10.0$ to introduce normalization errors. We use our method with the Stein variational gradient flow (\textsf{FM{-}Stein}) and compare it with the Fourier ergodic metric optimized (\textsf{Fourier}) and the kernelized ergodic metric optimized (\textsf{Kernel}), all three of which require access to the probability density function of the target distribution. We optimize the Fourier ergodic metric using the augmented Lagrange multiplier-based method from~\citet{dong_time-optimal_2024} and the kernelized ergodic metric using the projection-based trajectory optimization method from~\citet{sun_fast_2025}, as these implementations yield the best empirical results and thus better demonstrate the properties of each ergodic metric. Each method is allowed to optimize for up to $0.5$s wall clock time in each trial, and we report the Fourier ergodic metric with respect to the correctly normalized target distribution for the converged trajectory. We use second-order point mass dynamics for all the tests.

\begin{figure*}[h!]
    \centering
    \begin{subfigure}[b]{\textwidth}
        \centering
        \includegraphics[width=0.99\textwidth]{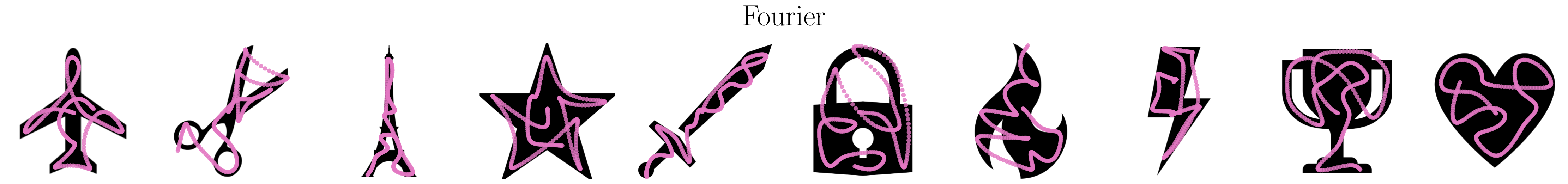}
    \end{subfigure}
    \vspace{+0.5em}
    \begin{subfigure}[b]{\textwidth}
        \centering
        \includegraphics[width=0.99\textwidth]{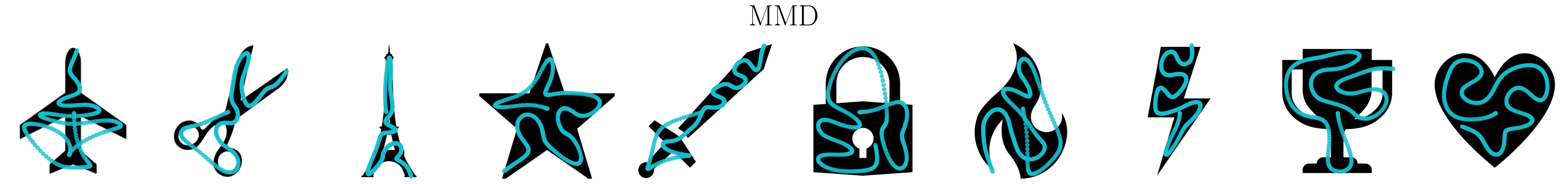}
    \end{subfigure}
    \vspace{+0.5em}
    \begin{subfigure}[b]{\textwidth}
        \centering
        \includegraphics[width=0.99\textwidth]{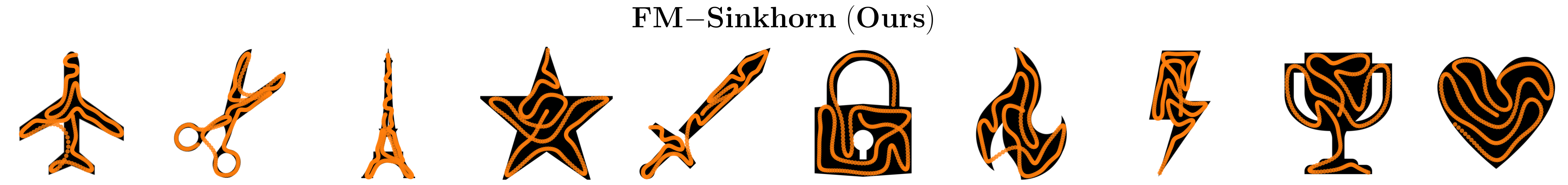}
    \end{subfigure} 
    \vspace{-1em}
    \caption{\textbf{Qualitative results for Q3.A}. Our method better captures the non-smooth geometry of the target distributions by leveraging the Sinkhorn divergence flow.}
    \label{fig:nonsmooth_trajs}
\end{figure*}

\begin{figure*}[htbp]
    \centering
    \includegraphics[width=\textwidth]{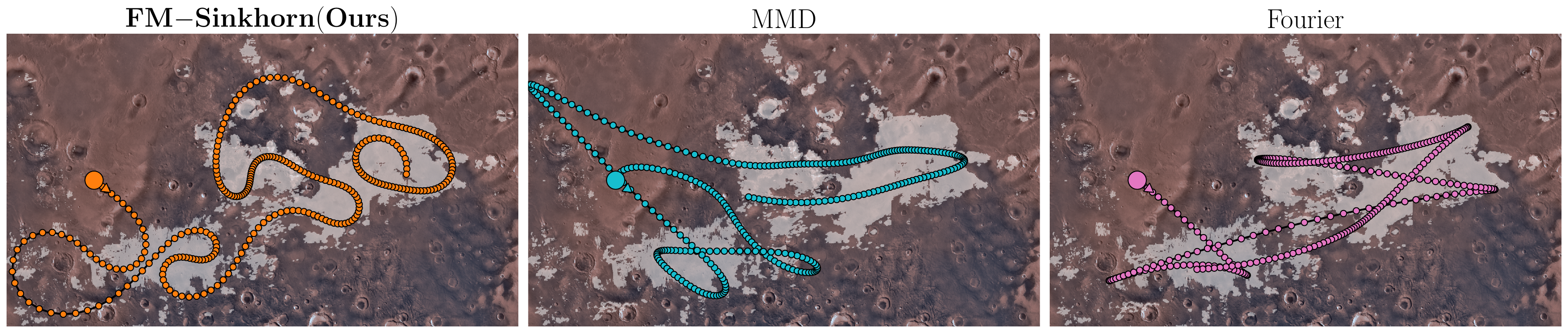}
    \caption{\textbf{Qualitative results for Q3.B}. The target distribution is shown in white color. Compared to MMD and Fourier ergodic metric, Sinkhorn divergence leads to better coverage trajectories on target distributions with irregular, discontinuous supports, which are commonly seen in real-world exploration problems.}
    \label{fig:mars_trajs}
    \vspace{-1em}
\end{figure*}

\begin{figure}[htbp]  
    \centering
    \includegraphics[width=0.8\columnwidth]{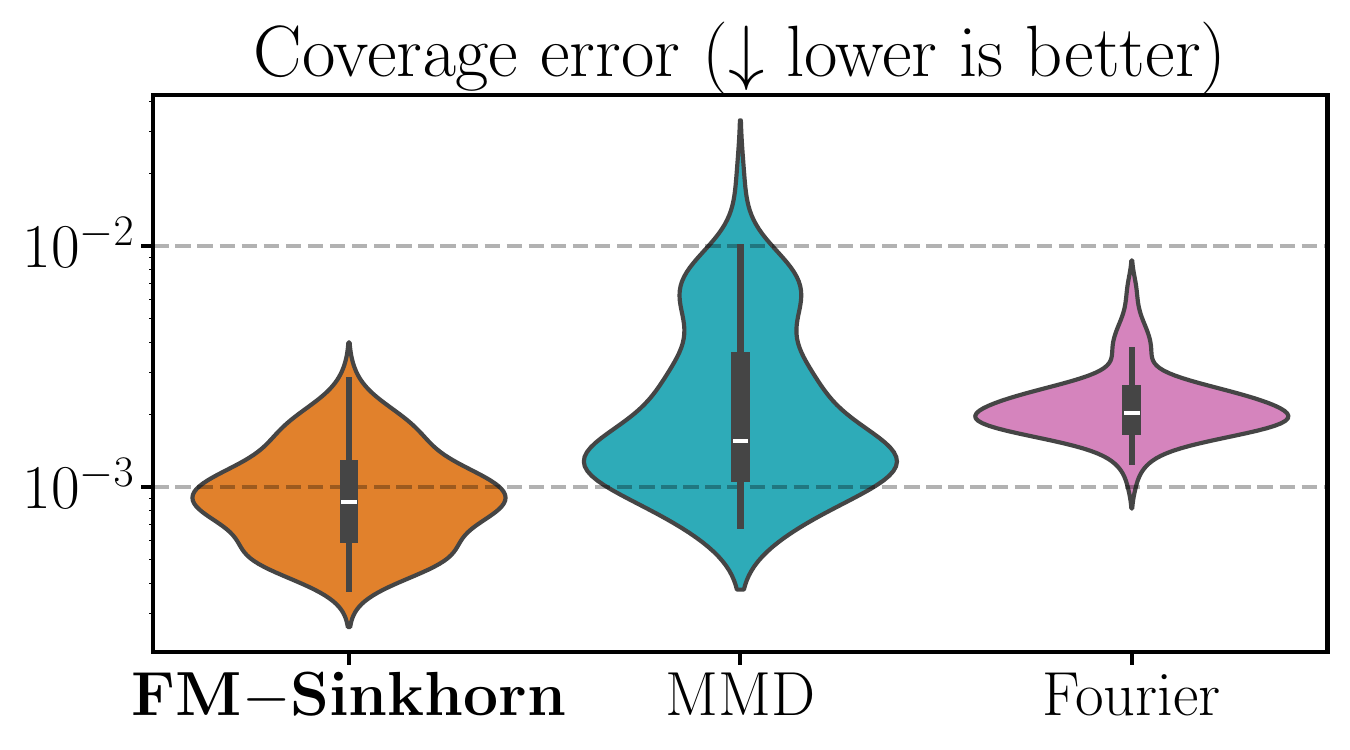} 
    \caption{\textbf{Quantitative results for Q3.A}. Our method consistently reaches lower coverage errors compared to the baselines. The white line in the violin plot is the median of the results.}
    \label{fig:nonsmooth_results}
    \vspace{-1em}
\end{figure}

\begin{figure}[htbp]  
    \centering
    \includegraphics[width=0.8\columnwidth]{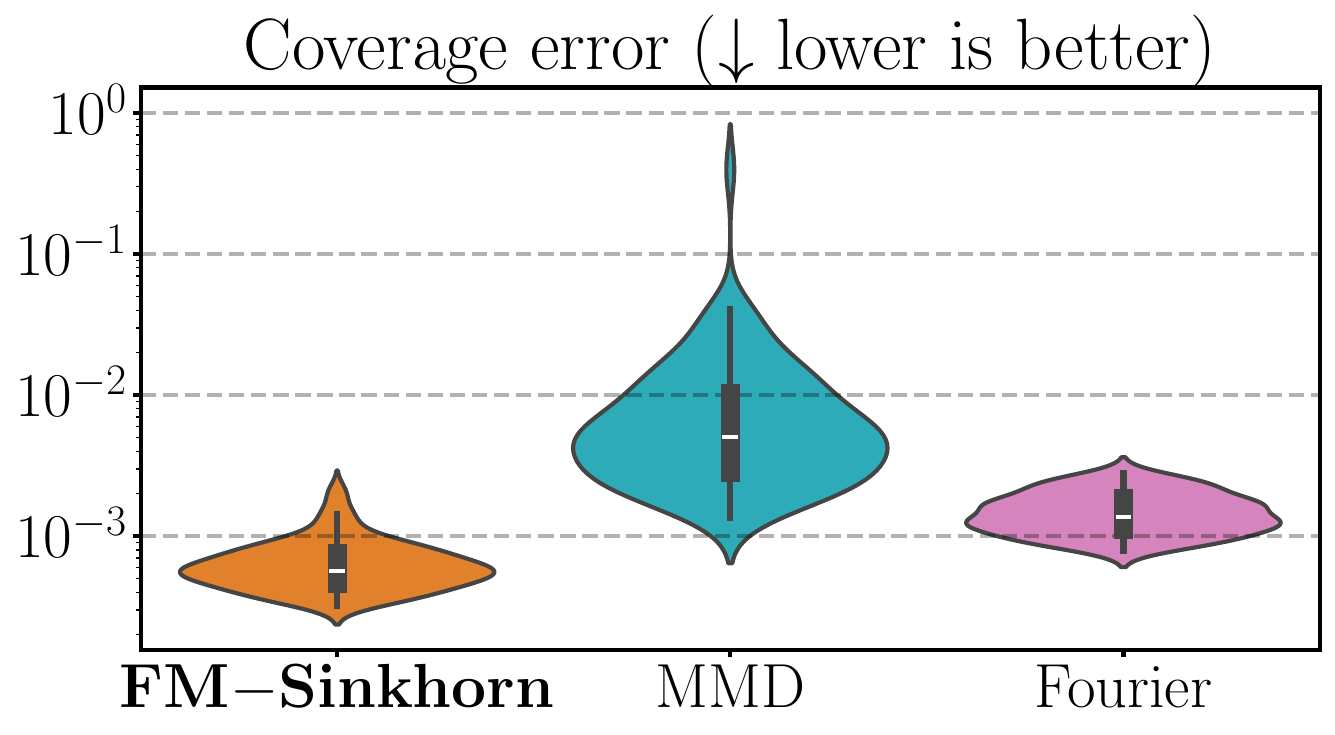} 
    \caption{\textbf{Quantitative results for Q3.B}. Our method consistently reaches lower coverage errors compared to the baselines. The white line in the violin plot is the median of the results.}
    \label{fig:mars_metrics}
    \vspace{-1em}
\end{figure}

\noindent\textbf{[Q2.A results] } The statistics of the ergodic metric at convergence for each method are shown in Fig.~\ref{fig:comparison_fourier_stein} (left). The results show that the ergodic metric value that our method converges to is, on average, one order of magnitude lower than the two baselines, despite the inaccurately normalized target distributions and all within the same amount of wall clock time. Compared to the Fourier metric, our method is also more consistent across the 100 trials. A qualitative comparison between the trajectories is also shown in Fig.~\ref{fig:comparison_fourier_stein} (right). The results show that both baselines fail to sufficiently explore the target distribution under normalization error, while our method successfully generates an accurate coverage trajectory.

\noindent\textbf{[Benchmark Q2.B design] } We test the three methods in Benchmark Q2.A in a simulated search and rescue (SAS) benchmark. Given a search space, we assume there are unknown regions within the space that emit positive life signals. A set of prior negative and positive life signals are sparsely sampled across the search space, the target distribution is the likelihood of measuring positive life signal across the search space computed using Gaussian processes (GPs) from the prior measurements, which is an unnormalized density function. The task of the benchmark is to explore the space to collect positive life signals given the GP estimation. We model the robot dynamics as a second-order differential-drive system. We conduct 100 test trials with randomized prior measurement and robot initial state. 

\noindent\textbf{[Q2.B results] } The statistics of the number of positive life signals collected by each method are shown in Fig.~\ref{fig:comparison_sns} (left), where our method consistently collect more positive life signals compared to the two baseline SOTA methods. Furthermore, the coverage trajectories from the tested methods in a representative trial are shown in Fig.~\ref{fig:comparison_sns} (right). These results show the advantage of the Stein variational gradient flow in practical applications, where unnormalized target distributions negatively affect the performance of existing ergodic coverage methods but does not affect our flow matching method.

\begin{figure*}[t!] 
    \centering    
    \includegraphics[width=\linewidth, keepaspectratio]{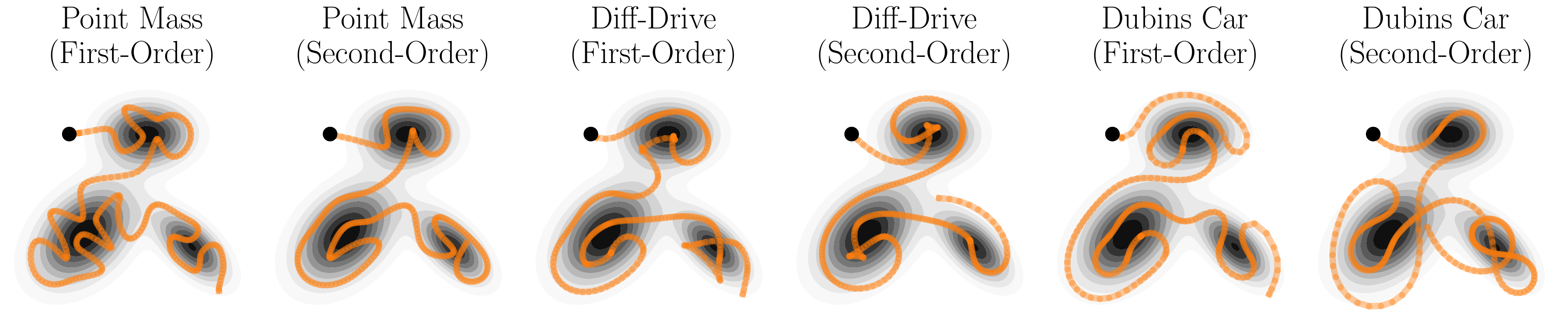} 
    \vspace{-1em}
    \caption{\textbf{Results for Q4}. Our method generates comparable ergodic coverage trajectories across six different dynamics.}
    \label{fig:compare_dynamics}
\end{figure*}

\begin{figure}[t!] 
    \centering    
    \includegraphics[width=0.49\textwidth, keepaspectratio]{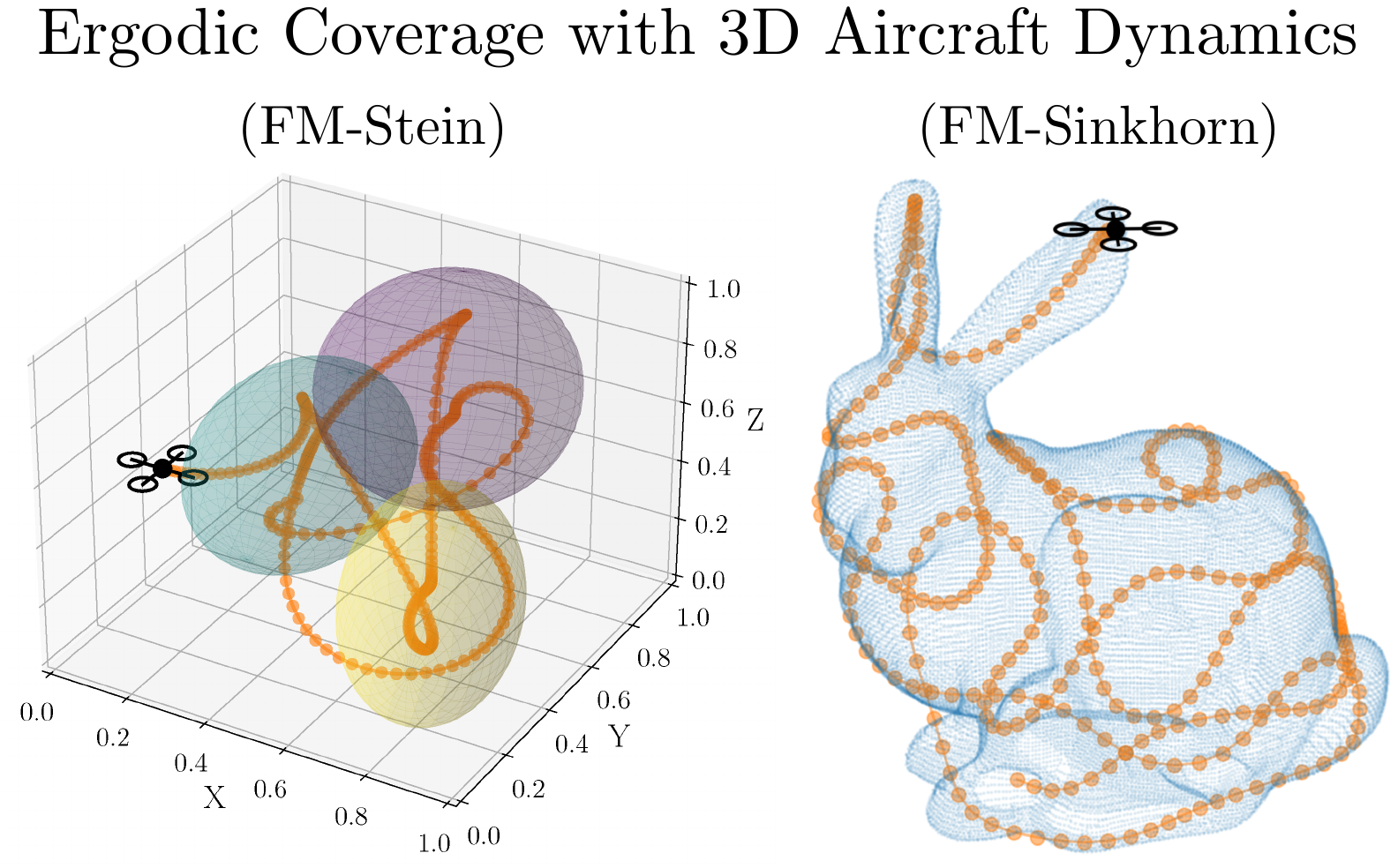} 
    \vspace{-1em}
    \caption{\textbf{Results for Q4}. Qualitative results with 3D aircraft dynamics using both the Stein variational gradient flow and Sinkhorn divergence flow.}
    \label{fig:3d_trajs}
    \vspace{-1em}
\end{figure}

\subsection{Experiment results for Q3}

We answer Q3 with two numerical benchmarks (Q3.A and Q3.B). Details of benchmark design and results are as follow.

\noindent\textbf{[Benchmark Q3.A design] } We choose 10 open source stock icons as the non-smooth target distributions (shown in Fig.~\ref{fig:nonsmooth_trajs}) and conduct 10 trials with randomized initial positions on each icon for all the methods tested (100 trials in total). We use our method with the Sinkhorn divergence flow (\textsf{FM{-}Sinkhorn}) and compare it to the same Fourier ergodic metric baseline from Benchmark Q3 (\textsf{Fourier}) and the maximum mean discrepancy metric optimized using the same augmented Lagrange multiplier-based method~\cite{hughes_ergodic_2024} (\textsf{MMD}). The target distribution is represented as discrete grids for \textsf{Fourier} and discrete samples (drawn using rejection sampling) for \textsf{FM{-}Sinkhorn} and \textsf{MMD}. Since the target distributions are non-smooth uniform distributions, we measure the coverage error using the trajectory uniformity metric in~\citet{mathew_metrics_2011} (equation 4). We use second-order point mass dynamics for all the tests. 

\noindent\textbf{[Q3.A results] } In Fig.~\ref{fig:nonsmooth_trajs}, we qualitatively show trajectories for each method across the 10 tested target distributions. Our method better captures the non-smooth geometry of the target distributions by leveraging the Sinkhorn divergence flow, which is a known advantage of optimal transport metrics~\cite{peyre_computational_2019}. The quantitative results are shown in Fig.~\ref{fig:nonsmooth_results}. On average, our method has a coverage error of $0.001$, which is two times lower than \textsf{MMD} (0.0028) and \textsf{Fourier} (0.0023).

\noindent\textbf{[Benchmark Q3.B design] } We further test the methods from Benchmark Q3.A using the NASA Mars Water Resource Maps (MWR)\footnote{\url{https://ammos.nasa.gov/marswatermaps/?mission=MWR}} as the target distribution. We model the robot dynamics as a second-order differential-drive system. We conduct 100 test trials with randomized robot initial state and evaluate the coverage error using the same metric from Benchmark Q3.A.

\noindent\textbf{[Q3.B results] } In Fig.~\ref{fig:mars_trajs}, we qualitatively show the trajectories from a representative trial. Quantitative results are shown in Fig.~\ref{fig:mars_metrics}. Different from the target distributions in Benchmark Q3.A, the target distributions from the Mars Water Resource Maps has irregular and discontinuous supports, which are common in real-world exploration tasks and create extra challenges for ergodic coverage. From the results, we can see that ergodic coverage based on MMD is particularly affected by such target distributions, while our method using the Sinkhorn divergence flow consistently produces the best coverage trajectories across the benchmark trials, demonstrating the advantage of optimal transport-based ergodic metric in practical applications.

\begin{figure}[t!] 
    \centering    
    \includegraphics[width=0.49\textwidth, keepaspectratio]{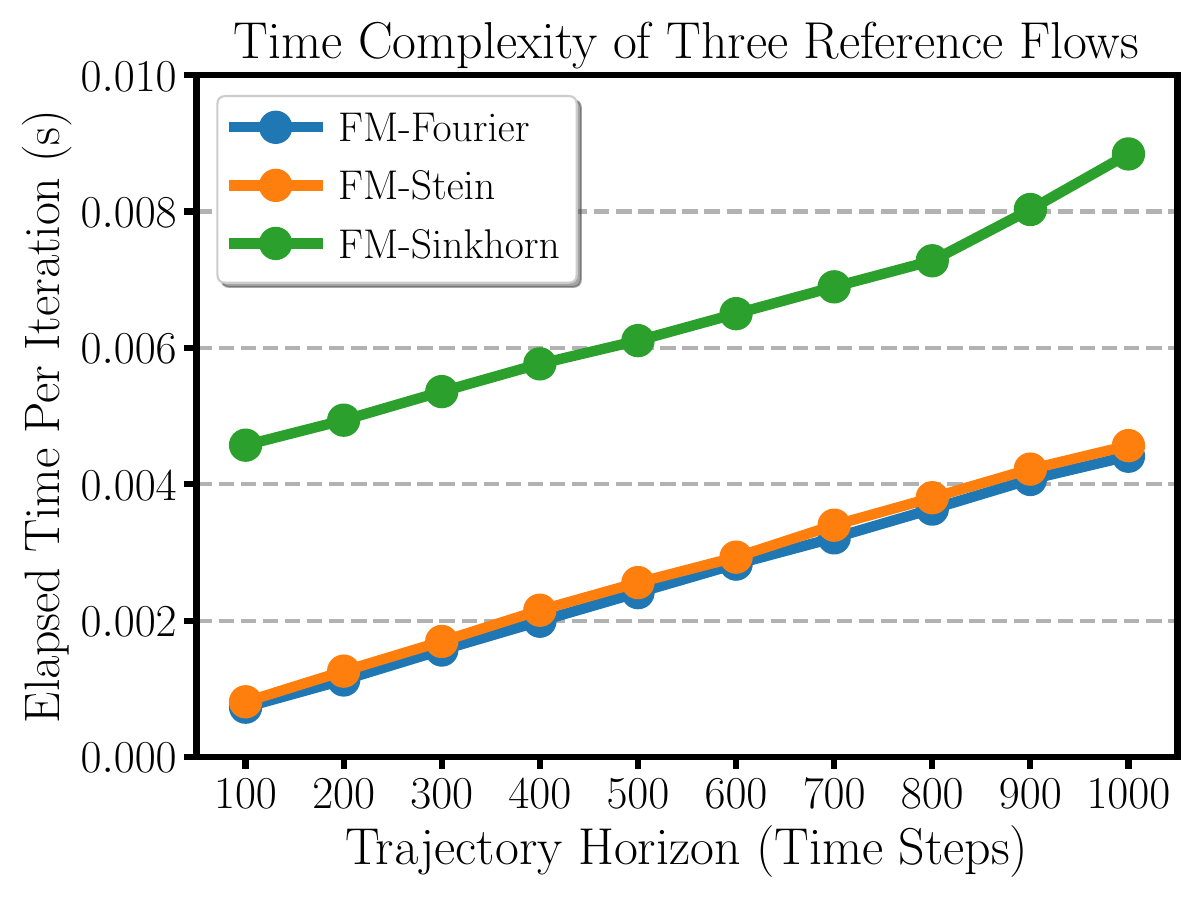} 
    \vspace{-1em}
    \caption{\textbf{Results for Q5}. Elapsed time of solving the linear quadratic flow matching problem across different trajectory horizons with all three reference flows.}
    \label{fig:flow_complexity}
    \vspace{-1em}
\end{figure}

\subsection{Experiment results for Q4}

We specify the target distribution as a trimodal Gaussian mixture model and use our method with the Stein variational gradient flow. We fix the initial position of the robot and test our method with six different dynamics: first and second-order point mass dynamics, first and second-order differential drive dynamics, and first and second-order Dubins car dynamics. The results are qualitatively shown in Fig.~\ref{fig:compare_dynamics}, which show that our method generates comparable coverage trajectories across different linear and nonlinear dynamics.

\begin{figure*}[t!]
    \centering
    \begin{subfigure}[b]{\textwidth}
        \centering
        \includegraphics[width=0.96\textwidth]{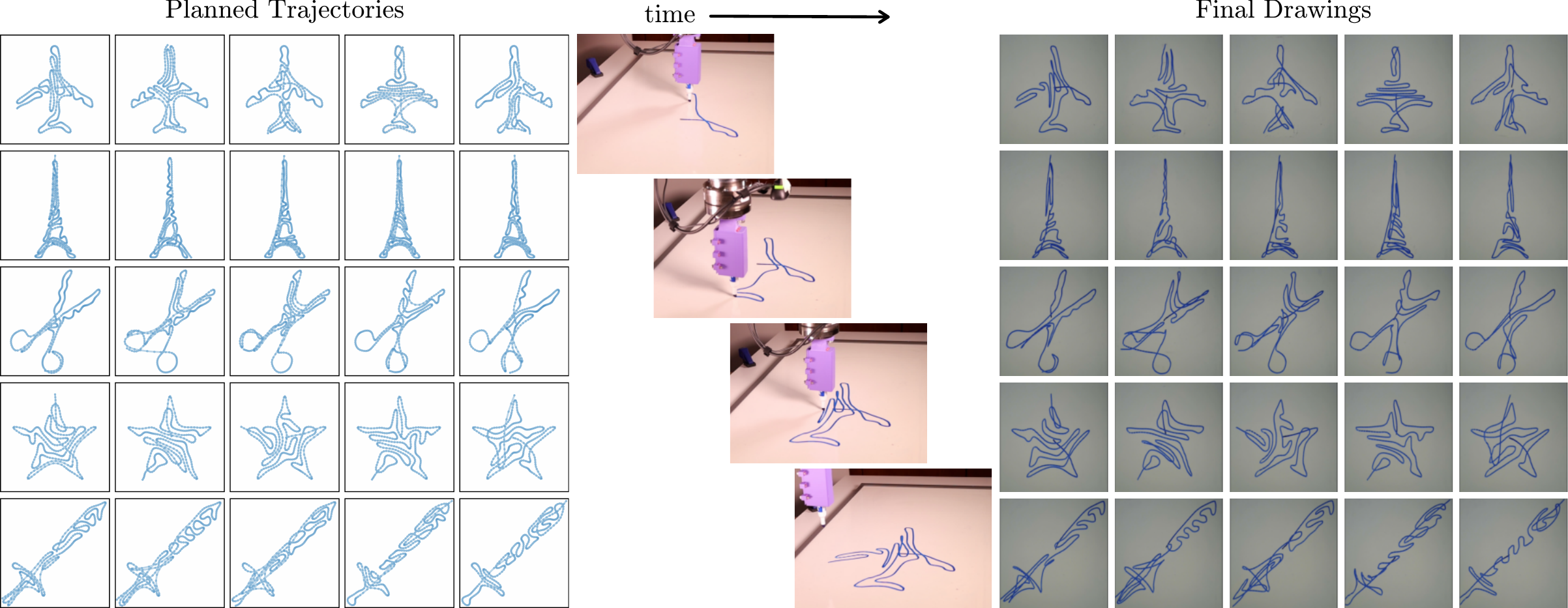}
    \end{subfigure}    
    \vspace{-1em}
    \caption{\textbf{Drawing task.} Planned object drawing trajectories (left), an example time sequence (middle), and final robot drawings (right). See videos on project website: \url{https://murpheylab.github.io/lqr-flow-matching/}.}
    \label{fig:sequence_drawing}
    \vspace{-1em}
\end{figure*}

Furthermore, we qualitatively test our method for 3D coverage tasks with the 3D aircraft dynamics from~\citet{lee_stein_2024}. Results from both the Stein variational gradient flow and Sinkhorn divergence flow are shown in Fig.~\ref{fig:3d_trajs} and are included in our project website.

\subsection{Experiment results for Q5}

We evaluate the elapsed time of solving the linear quadratic flow matching problem (\ref{eq:lq_flow_matching}) across different trajectory time horizons (100 to 1000 time steps) for all three reference flows. From the results shown in Fig.~\ref{fig:flow_complexity}, we can see the flow matching formula exhibits a linear time complexity for all three reference flows. Furthermore, flow matching with the Fourier ergodic metric flow has near identical computation time with the Stein variational gradient flow. While flow matching with Sinkhorn divergence flow takes longer time, it is still sufficient for real-time trajectory optimization.

\section{Hardware Demonstrations}

\begin{figure*}[h!]
    \centering
    \begin{subfigure}[b]{\textwidth}
        \centering
        \includegraphics[width=0.96\textwidth]{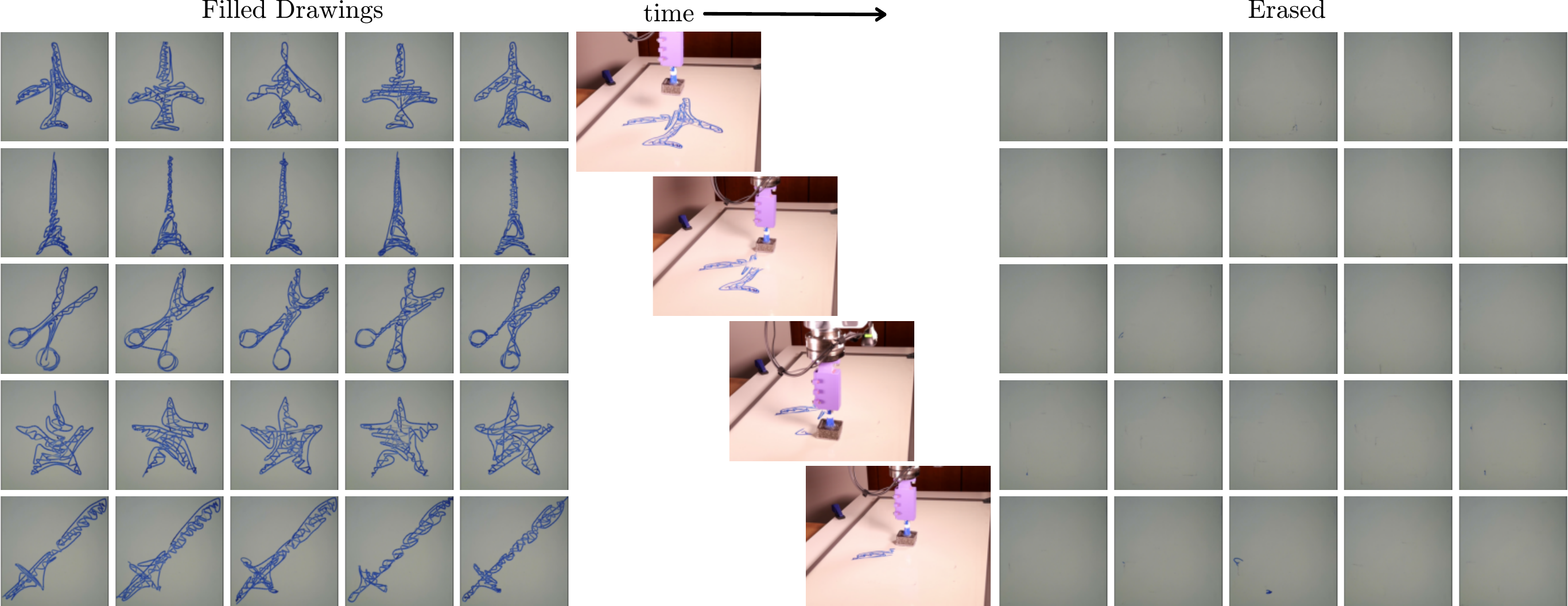}
    \end{subfigure}    
    \vspace{-1em}
    \caption{\textbf{Erasing task.} Initial filled object drawings to be erased (left), an example time sequence (middle), and final result for each erased object (right). See videos on project website: \url{https://murpheylab.github.io/lqr-flow-matching/}.}
    \label{fig:sequence_erasing}
    \vspace{-1em}
\end{figure*}

We demonstrate the effectiveness of our method on a Franka Emika Panda robot for a series of drawing and erasing tasks. For both tasks, we use the Sinkhorn divergence flow to generate ergodic coverage trajectories over five open source stock icons as the target distributions. For the drawing task (see Fig.~\ref{fig:sequence_drawing}), the robot end-effector is equipped with a dry-erase marker, which is used to draw the desired trajectories on a whiteboard. For the erasing task (see Fig.~\ref{fig:sequence_erasing}), we manually fill in regions of the whiteboard corresponding to the desired objects and attach an eraser to the robot's end-effector. For both hardware tasks, the trajectories are pre-planned in the x-y plane using double-integrator point mass dynamics and are executed open-loop. The robot controller operates at 3 Hz for drawing and 6 Hz for erasing. 

In Fig.~\ref{fig:sequence_drawing}, we qualitatively show that all planned drawing trajectories were feasible for the robot to execute. The Final drawings (right) closely resemble the planned trajectories (left). In Fig.~\ref{fig:sequence_erasing}, we qualitatively show that the robot successfully erased all filled object drawings. We further conduct closed-loop erasing experiments with the onboard camera of the robot providing image feedback, where new drawings are added during erasing. Videos of both the open-loop and closed-loop hardware demonstrations can be found on the project website, which illustrate the different paths executed for each object.

\section{Limitations}

The proposed flow matching framework focuses on trajectory optimization, aiming to generate the most ergodic trajectory within a fixed time horizon. In contrast, feedback control-based ergodic coverage methods~\cite{ivic_ergodicity-based_2017,shetty_ergodic_2022,bilaloglu_tactile_2025} offer improved computational efficiency and robustness to runtime disturbances, but typically require more time to achieve comparable coverage~\cite{sun_fast_2025}. The Fourier ergodic metric enables online model predictive control while still guaranteeing the asymptotic convergence~\cite{mathew_metrics_2011,mavrommati_real-time_2018}, such that the robot can rapidly react to unknown perturbations during the task (e.g., noisy odometry readings) while still maintaining good coverage performance. However, it is unclear if a model predictive control variant of the flow matching formula has a similar asymptotic convergence guarantee for the Stein variational gradient flow or the Sinkhorn divergence flow. As a result, all the experiments in this work are based on long-horizon trajectory optimization, which generates a reference trajectory prior to the task, and the control actions for the robot during runtime are generated through a lower-level tracking-based controller. While the flow matching method has sufficient computational efficiency to replan during runtime, the lack of a model predictive control or feedback control formula with asymptotic convergence guarantees remains a limitation of this work.

In addition, there are limitations for the two alternative ergodic metrics introduced in this work. Despite the ability to generate ergodic trajectories over unnormalized distributions using the Stein variational gradient flow, the score function in (\ref{eq:stein_flow}) cannot be computed in closed-form if the target distribution is represented as samples. While there exist techniques such as score matching~\cite{song_sliced_2020,vincent_connection_2011} to estimate the score function from data, the optimization cannot be conducted in real-time, and the estimated score function can be inaccurate in regions with low sample density~\cite{song_sliced_2020}. Furthermore, while the kernel function improves computational efficiency and flexibility, the kernel function only provides local correlation between the states; thus, it could lead to sub-optimal performance over spatially disconnected target distributions,  which is a common issue in Stein variational gradient flows~\cite{liu_stein_2016}. On the other hand, the Sinkhorn divergence flow is limited by the computational cost associated with evaluating the optimal transport distance---the computation increases linearly with the number of samples from the target distribution~\cite{genevay_sample_2019}, which could have a larger impact as the size of search space increases.

Lastly, while formal convergence analyses exist for standard trajectory optimization frameworks in ergodic coverage~\cite{miller_trajectory_2013,dong_time-optimal_2024}, such analysis is currently lacking for our flow matching-based ergodic coverage framework. As a result, the step size parameters $\eta$, $\Delta \tau$, and the number of iterations in Algorithm~\ref{algo:ergodic_coverage} need be manually tuned in practice.

\section{Conclusions}

In this work, we introduce a novel optimization paradigm for ergodic coverage based on flow matching. Our method enables alternative ergodic metrics previously infeasible for control synthesis to overcome the limitations of existing metrics with no computational overhead. We show that using the Stein variational gradient flow improves the robustness of ergodic coverage on unnormalized target distributions, and using the Sinkhorn divergence flow improves the ergodic coverage performance over non-smooth distributions with irregular supports. We further validate the effectiveness of our method through hardware demonstrations on a Franka robot.

Our work provides a new perspective for bridging statistical inference and the decision-making of embodied agents, where the motion of the agents serves as a dynamically-constrained inferred posterior. While this work focuses on two specific methods to construct the flow, recent advances in flow-based inference and generative models suggest that the methods in this paper can be paired with any statistical inference framework that admits flow-based inference, providing opportunities for advances in embodied intelligence. The coupling between statistical inference and motion synthesis in our work could further enable the integration of ergodic coverage as a behavior characterization framework, which extends the conventional notion of trajectory tracking to distribution tracking, into learning-based methods for motion synthesis or policy representations.

\section*{Acknowledgments}

This work is supported by ARO grant W911NF-19-1-0233 and W911NF-22-1-0286. The views expressed are the authors' and not necessarily those of the funders.

\bibliographystyle{plainnat_titlelink}
\bibliography{references}

\end{document}